\def\year{2022}\relax
\documentclass[letterpaper]{article} % DO NOT CHANGE THIS
\usepackage{aaai2026}   % DO NOT CHANGE THIS
\usepackage{times}  % DO NOT CHANGE THIS
\usepackage{helvet}  % DO NOT CHANGE THIS
\usepackage{courier}  % DO NOT CHANGE THIS
\usepackage[hyphens]{url}  % DO NOT CHANGE THIS
\usepackage{graphicx} % DO NOT CHANGE THIS
\usepackage{amssymb}
\usepackage{amsmath}
\usepackage{mathtools}
\usepackage{tabularx}
\usepackage{booktabs}
\usepackage{multirow}
\usepackage{xcolor}
\usepackage{float}

\usepackage{natbib}  % DO NOT CHANGE THIS AND DO NOT ADD ANY OPTIONS TO IT
\usepackage{caption} % DO NOT CHANGE THIS AND DO NOT ADD ANY OPTIONS TO IT
\DeclareCaptionStyle{ruled}{labelfont=normalfont,labelsep=colon,strut=off} % DO NOT CHANGE THIS
\usepackage{algorithm}
\usepackage{algorithmic}
\usepackage{subcaption}

\usepackage{newfloat}
\usepackage{listings}
\floatstyle{ruled}
\newfloat{listing}{tb}{lst}{}
\floatname{listing}{Listing}
\title{Towards Safer Social Media Platforms: Performant Few-Shot Harm Detection using Large Language Models
}

\author{
    Akash Bonagiri\textsuperscript{\rm 1}\equalcontrib,
    Lucen Li\textsuperscript{\rm 1}\equalcontrib,
    Rajvardhan Oak\textsuperscript{\rm 1,3}\equalcontrib,
    Zeerak Babar\textsuperscript{\rm 1},\\
    Magdalena Wojcieszak\textsuperscript{\rm 1},
    Anshuman Chhabra\textsuperscript{\rm 2}
}
\affiliations {
    \textsuperscript{\rm 1}University of California, Davis\\
    \textsuperscript{\rm 2}University of South Florida\\
    \textsuperscript{\rm 3}Meta\\
    \textit{\{sbonagiri, lcnli, rvoak, zebabar, mwojcieszak\}@ucdavis.edu}, \textit{anshumanc@usf.edu}
    
}

\begin{document}

\maketitle

\noindent \textcolor{red}{Note: \textit{This paper may contain examples of content that is inappropriate, offensive or amounting to hate speech.}\\}

\begin{abstract}
%Write abstract after everything else is done
The prevalence of harmful content on social media platforms poses significant risks to users and society, necessitating more effective and scalable content moderation strategies. Current approaches rely on human moderators, supervised classifiers, and large volumes of training data, and often struggle with scalability, subjectivity, and the dynamic nature of harmful content (e.g., violent content, dangerous challenge trends, etc.). To bridge these gaps, we utilize Large Language Models (LLMs) to undertake few-shot dynamic content moderation via in-context learning. Through extensive experiments on multiple LLMs, we demonstrate that our few-shot approaches can outperform existing proprietary baselines (Perspective and OpenAI Moderation) as well as prior state-of-the-art few-shot learning methods, in identifying harm. We also incorporate visual information (video thumbnails) and assess if different multimodal techniques improve model performance. Our results underscore the significant benefits of employing LLM based methods for scalable and dynamic harmful content moderation online.
\end{abstract}

\section{Introduction}
Social media platforms are an integral part of people's daily lives, influencing how individuals communicate, share information, and connect with others. Platforms such as Facebook and YouTube are not only tools for interaction, but also play a crucial role in networking, marketing, education, sales, and political campaigns.
However, platforms can also inadvertently disseminate a myriad of harmful content to their users and influence key decisions \cite{badam2022aletheia}. Over 30\% of users have encountered hate speech or aggressive behavior online\footnote{{\url{https://www.pewresearch.org/internet/2021/01/13/the-state-of-online-harassment/}.}}, false and misleading content frequently spreads faster than truthful information online \cite{aimeur2023fake}, vulnerable users are targeted with recommendations to problematic and distressing content \cite{hilbert2023bigtech}, and dangerous challenges are propagated through social media ~\cite{haidt2023social}. Exposure to harmful content can have negative consequences for individuals, groups, and the society at large \cite{haidt2023social}.

\begin{figure*}[ht]
  \centering
  \includegraphics[width=0.92\textwidth]{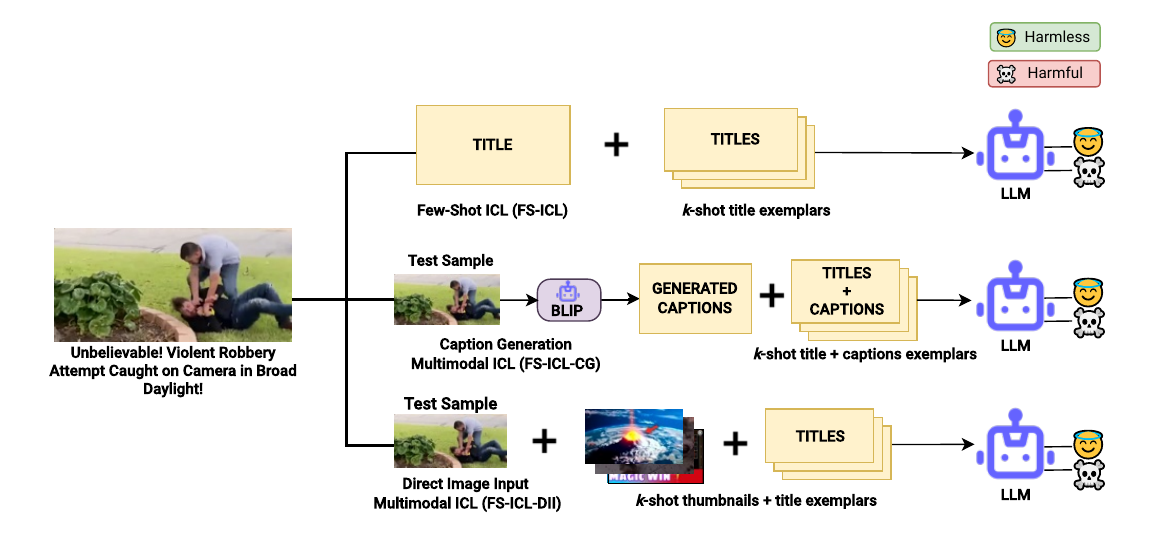}\vspace{-3mm}
  \caption{Our overall experimental framework utilizing LLMs for $k$-shot harmful content classification. The top approach showcases few-shot text-only in-context learning using the video's title (FS-ICL). The other two methods (FS-ICL-CG) are multimodal and augment the video title information with visual input (e.g. video thumbnail). In FS-ICL-CG we utilize a caption generation module (BLIP) to convert the visual input into text and feed these to a text-only LLM. In FS-ICL-DII we utilize fully multimodal LLMs and feed the visual input to the model directly. The output of each method is a classification of \textit{Harmful/Harmless} for each given test input. Note that by setting $k=0$ we can obtain zero-shot learning (ZSL) variants for each of the aforementioned approaches.}\label{fig:intro}\vspace{-3mm}
\end{figure*}

Given these potential consequences, platforms take a number of steps to decrease exposure to harmful content. These approaches range from removal (e.g., banning content that encourages violence, suicide, or eating disorders~\cite{youtubetos}), limiting amplification without removal (e.g., minimizing the sharing of controversial posts), or providing additional information alongside potentially harmful content (e.g., offering links to fact-checking pages to debunk false claims~\cite{twittercommunity}). All these strategies depend on dynamic and scalable approaches for identifying ever-changing harmful content online. 

To this end, platforms generally employ a combination of human moderators and automated classifiers~\cite{facebook_help_moderation}. Human moderators review content flagged as harmful either by users or automated content moderation systems, and decide if it violates the platform's policies. In some cases, Machine Learning (ML) classifiers trained on large human-annotated datasets~\cite{perspective_api, openai_moderation} are used to identify misinformation, hate speech, sexually explicit material, among other types of harm \cite{akash2021poster}. However, both human and supervised content moderation classifiers possess significant drawbacks. Human moderation is subjective, not easily scalable, and prone to operator error~\cite{gillespie2020content}. In turn, supervised ML classifiers are inefficient, as they require large volumes of human-annotated data for training and struggle with nuances in language such as sarcasm and context-specific meanings~\cite{baruah2020context}. These issues with supervised ML models are further compounded by the fact that harmful content classification is a \textit{dynamic temporal problem} (e.g., new conspiracies or new dangerous challenges become popular). As a result, content moderation classifiers are susceptible to concept drift~\cite{quinonero2022dataset} and require humans to annotate large amounts of data at frequent intervals, exacerbating the aforementioned scalability issues.

\looseness-1 It is apparent then that an \textit{ideal} solution to the harmful content moderation problem needs to be \textit{highly scalable} (i.e. require minimal supervised signal and human effort) and \textit{highly effective} (i.e. attain human moderator-level accuracy). Moreover, the approach should be easily extensible and adapted to new types or categories of harm as they emerge online. In this paper, we demonstrate that this is indeed possible with Large Language Models (LLMs). We utilize a recent multimodal dataset~\cite{Jo_2024} consisting of 19,422 YouTube videos, annotated by domain experts, Amazon MTurkers, and LLMs for six different harm categories: \textit{Information, Hate and Harassment, Addictive, Clickbait, Sexual} and \textit{Physical Harms}.

\looseness-1 Our overall experimental framework is outlined in Figure \ref{fig:intro}. We consider a number of \textit{open-source} and \textit{closed-source} LLMs\footnote{We primarily consider Llama2-12B, Mistral-7B, GPT-4o-Mini, and GPT-3.5-Turbo in this work.} to first show that LLMs can classify harmful content even in the zero-shot setting (i.e. no additional information is provided to them except for the task definition) better than the currently used proprietary baselines, such as Google's Perspective API~\cite{perspective_api} and OpenAI's Moderation API~\cite{openai_moderation}. We then show that LLM performance for social media content moderation can be significantly improved by using \textit{state-of-the-art few-shot in-context learning} \cite{dong2022survey, gupta2023coverage} approaches, where a few examples demonstrating harmful/harmless content are provided to the model in the instruction prompt. We showcase how these methods outperform existing deep learning few-shot classification models, establishing LLMs as the state-of-the-art in classifying social media harm in the minimal supervision setting. Finally, we investigate whether LLM performance at this task can be further augmented by incorporating visual input in the learning pipeline (i.e. video thumbnails) and employing Vision-Language Models (VLMs). Our results demonstrate important future directions for social media harm classification and content moderation in the era of LLMs, and pave the way for accelerated efforts in this promising research landscape. 

In sum, we make the following contributions:

\begin{itemize}%[wide]
    \item We show that LLMs, even in the zero-shot setting, can outperform existing proprietary classifiers for harm identification, such as Perspective API and OpenAI Moderation API.
    \item We then examine the performance of LLMs for harm detection under the few-shot setting and showcase that our approaches can adapt to the dynamic nature of harmful content with as few as 8 exemplars. We employ state-of-the-art inference-time approaches for in-context learning to select exemplars and find that LLMs can outperform existing few-shot learning methods when provided the same number of exemplars.
    \item We also incorporate multimodal LLMs and visual input (video thumbnails) to analyze content, and find that this leads to improved accuracy in identifying harmful content. However, these performance gains are contingent on the LLM being used and open-source multimodal models (e.g., LLaVa \cite{liu2023llava}) exhibit lower performance compared to their closed-source counterparts (e.g. GPT-4o \cite{OpenAI_GPT-4o_mini}).
\end{itemize}

\section{Related Work}

\looseness-1\textbf{Categorizing Social Media Harms.}
Social media platforms host a range of content that can be categorized as harmful.
%
%@@@
 The combination of platform affordances, which make (problematic) content production and dissemination cheap and efficient \cite{cinelli2021echo}, and recommender systems trained to maximize user engagement, make exposure to online harm far from infrequent. Digital traces each user leaves on platforms reveal information about the user’s emotions \cite{gui2022emotion}, personality \cite{wang2023personality}, substance use \cite{coppersmith2021substance}, and sexual orientation \cite{wang2023sexual}. In their effort to maximize engagement, algorithms can capitalize on this inferred information to recommend content that can inadvertently expose users to various harms (e.g., addictive content to users known to use substances, suicidal content to depressed users, misinformation to users interested in herbology) \cite{montag2021detrimental, tiktok21}. Meta-reviews have shown that 8\%-10\% of recommendations on platforms pose risks to users ~\cite{hilbert20248} and algorithmic audits have detected discriminatory or otherwise harmful biases in algorithms \cite{bandy2021problematic, haroon2023auditing}. For instance, distressing content is recommended to struggling adolescents on YouTube and Instagram \cite{hilbert2023bigtech} and up to 40\% of health-related content on social media can be classified as misleading or false \cite{cinelli2020covid}. Exposure to various categories of online harm can decrease mental health, foment addictions, or even lead to physical harms \cite{haidt2023social, Surgeon}.

%Indeed, whistleblowers reveal that platforms are aware of the harms they inflict \cite{haugen2023power}
%\cite{marwick2017media} provided a foundational framework categorizing online harm into several distinct types, including harassment and disinformation, which they linked to broader sociopolitical impacts. \cite{cinelli2020covid} offer insights into the scale of misinformation, noting that up to 40\% of content related to health topics on social media can be classified as misleading or false.
%Several works~\cite{rossi2021closed, pariser2011filter} have shown that recommendation algorithms on social media create \textit{filter bubbles} which can cause echo chambers and the rapid spread of misinformation~\cite{tomlein2021audit}. A meta-analysis~\cite{hilbert20248} shows that 8\%-10\% of recommendations pose detectable risks to users and that 20\% of adolescents report harm from social media use. 
% Do we need this here? We cover why Claire's dataset in 4.1
%\raj{Need 1-2 lines on why Claire's dataset and harm taxonomy improves upon (or at least adheres to) these existing harm categories/research which is why we use this dataset in our work.}

\vspace{1mm}\noindent\textbf{Automated Harm/Content Moderation.}
Given their detrimental effects, mitigating harms on social media has received widespread attention from the community. 
%
%@@@
Extensive research has been conducted on automated methods for detecting hate speech online \cite{jahan2023systematic} as well as misinformation \cite{mohsen2024automated}. For instance, \cite{dacon2022detecting} presented a BERT-based approach that demonstrated strong performance in recognizing hate speech, offensive language, and other harmful behaviors; \cite{song2020} proposed a multimodal stacking scheme that combines both visual (convolutional neural networks) and auditory (bi-directional recurrent neural networks) classifiers to detect online pornographic content, thereby improving accuracy.

\looseness-1 The advent of LLMs has led to new efforts in this space. The ability of LLMs to leverage learned patterns from large-scale data enables them to understand and generate coherent, contextually appropriate text. LLMs are highly effective at a variety of NLP tasks. They can operate in a zero-shot or few-shot setting and demonstrate understanding of context and user intentions without extensive task-specific training~\cite{brown2020language}. Accordingly, growing work has used LLMs to tackle online harms. For instance, \cite{wang2024intent} present LLM-based approach to generate counterspeech to mitigate hate speech. LLMs have also shown promise in other harm-related tasks, such as detecting clickbait~\cite{sekharan2023fine}, misinformation~\cite{santra2024leveraging} and abuse~\cite{nguyen2023fine}. 
While extensive, there are a few issues with extant work. First, each work focuses on one specific kind of harm and relies on data labeled for that harm category. As a result, research in this area is highly fragmented and there is a lack of solutions that approach this problem from a holistic perspective~\cite{arora2021, Jo_2024}.
Additionally, harm is an ever-evolving concept that changes as online communities and content adapt with time~\cite{mundriyevskaya2023online}~\cite{piot2025towards}. Existing approaches do not consider the concept drift and may not generalize to new kinds of harms without explicit labeled examples (which are expensive in terms of manual moderation efforts and time). 
We aim to address both of these challenges by presenting a harm-agnostic approach that does not require extensive labeling.

\section{LLMs for Harm Classification}
We now describe our LLM based strategies for harmful content moderation in the zero-shot and few-shot (in-context learning or ICL) unimodal (text-only) and multimodal (text + vision) settings. For undertaking multimodal classification, we describe two approaches: (1) converting the visual input into text for use with text-only LLMs (as in past work \cite{yang2024exploring}) and (2) directly utilizing visual input with natively multimodal LLMs/VLMs. We first define the harm classification problem analytically and then formalize our zero-shot and few-shot/ICL approaches. 

\subsection{Problem Formulation}
Let $X = \{x_i \mid x_i=<t_i, v_i>\}_{i=1}^n$ be a sequence of $n$ content instances, where each $x_i$ may contain both visual and textual components $v_i$ and $t_i$ respectively. Let $\mathcal{L}$ denote an LLM. Our goal is to learn a function \( f: X \rightarrow \{0,1\} \) using $\mathcal{L}$ that operates on a given prompt $\mathcal{P}$ to map every content instance $x \in X$ to a binary label, such that:
% \begin{center}
%     $
% f(x) =
% \begin{cases} 
% 1 & \text{if } x \text{ is harmful}, \\
% 0 & \text{if } x \text{ is harmless}.
% \end{cases}
% $
% \end{center}
\[
f(x) =
\begin{cases} 
1 & \text{if } x \text{ is harmful}, \\
0 & \text{if } x \text{ is harmless}.
\end{cases}
\]

\subsection{Zero-Shot Learning (ZSL)}
ZSL~\cite{yin2019benchmarking}~\cite{plaza2023respectful} is the the ability of a model to classify data it has never encountered during training and perform tasks without explicit pretraining. Because LLMs are trained on vast amounts of data, we expect them to have an inherent understanding of what constitutes harmful content. In the ZSL approach, we provide a sample as input to the model, and ask the LLM to classify it as \textit{harmful} or \textit{harmless}, without explicitly providing any demonstrations for what constitutes harmful/harmless.
Now, the function $f$ is defined using the LLM $\mathcal{L}$ and the content sample $x_i$, as:
% \begin{center}
%     $
% f_{\text{ZSL}}(x_i) \coloneq \mathcal{L}(t_i, \mathcal{P}_{\text{ZSL}}),
% $
% \end{center}
\[
f_{\text{ZSL}}(x_i) := \mathcal{L}(t_i, \mathcal{P}_{\text{ZSL}}),
\]

\noindent where $\mathcal{P}_{\text{ZSL}}$ is a prompt that guides $\mathcal{L}$ to classify $x_i$ as harmful or harmless without explicit examples of harmful content. The text input $t_i$ for each content sample $x_i$ comprises of the YouTube video's title.
The exact prompt we use can be found in Appendix~\ref{appendix:zslprompt}. 
%
% \noindent Similarly in the case of ZSL-CG, we pass images through BLIP model to generate captions. We augment these captions with the title (more detailed description in section 3.4) and follow ZSL approach mentioned above which we term as ZSL-CG. Now, the function $f$ is defined as $\mathcal{L}$, the content sample $x_i$, and denoting the captioning VLM as $\mathcal{V}$, the function $f$ incorporates a caption \( c_i \) generated by \( V \) from the visual component, i.e. $c_i = V(v_i)$, as follows:
% % \begin{center}
% \[
% f_{\text{ZSL-CG}}(x_i) \coloneq \mathcal{L}(t_i, c_i, \mathcal{P}_{\text{ZSL-CG}}).
% \]
% % \end{center}

% \noindent In the same way, for Direct Image Input (DII) approach which we term as ZSL-DII, we pass images directly as input to VLMs (more detailed description in  section 3.4). Now, the $f$ is defined as \[
%  f_{\text{ZSL-DII}}(x_i) \coloneq \mathcal{L}(v_i, t_i, \mathcal{P}_{\text{ZSL-DII}}).
% \]

\subsection{Few-Shot In-Context Learning (FS-ICL)}
Few-shot learning (FSL)~\cite{brown2020language} refers to the ability of a model to learn new tasks from a very limited amount of supervised data, often restricted to 8-14 annotated samples. In this setting, we utilize LLMs to perform harm classification by providing a small number of task-specific exemplars within the input prompt, via in-context learning \cite{dong2022survey} (ICL). We opt for coverage-based ICL approaches as they have been shown to be highly successful at generalizing to a wide array of domains \cite{gupta2023coverage}. These methods ensure maximally informative demonstrations are selected by submodular optimization, which efficiently maximizes the coverage of salient aspects of the content sample. Selectors such as BERTScore, cosine, and BM25 can guide the selection process \cite{chakraborti2024nlp4gov}. Thus, we utilize coverage-based approaches and augment the ZSL prompt to include $k$ labeled examples contained in the exemplar set $\mathcal{E} = \{(x_j, y_j)\}_{j=1}^{k}$, where $y_j \in \{0,1\}$ represents the true known label for $x_j$. The examples in $\mathcal{E}$ are selected based on their ability to maximize coverage of salient aspects and are ordered by relevance, with the most relevant examples placed closest to $x_i$.
The function $f$ takes the form:
% \begin{center}
%     $
%     f_{\text{FS-ICL}}(x_i) \coloneq \mathcal{L}(t_i, \mathcal{E}, \mathcal{P}_{\text{FS-ICL}}).
%     $
% \end{center}

\[
f_{\text{FS-ICL}}(x_i) \coloneq \mathcal{L}(t_i, \mathcal{E}, \mathcal{P}_{\text{FS-ICL}})
\]

\noindent Here, $\mathcal{P}_{\text{FS-ICL}}$ is an augmented prompt that instructs the LLM \( \mathcal{L} \) to utilize the demonstration set $\mathcal{E}$ while performing the classification based on the in-context exemplars utilizing BERTScore, Cosine, BM25 selectors. More details on selectors are provided in Appendix \ref{appendix:selectors}.

%Drawing from prior work, we experimented with four techniques for selecting $\mathcal{E}$ to be provided: BERTScore, BM25, cosine similarity and re-ordered cosine similarity.

\subsection{Multimodal FS-ICL}
\looseness-1 We enhance the FSL approach by including features from another modality (vision) along with the textual input for each content sample. For YouTube videos, this constitutes using video thumbnails. To incorporate visual input, we use multimodal LLMs (also referred to as VLMs): GPT-4o-Mini \cite{OpenAI_GPT-4o_mini}, OpenFlamingo \cite{Alayrac2022FlamingoAV} and LLaVA \cite{liu2023llava}. We employ models in two key ways: (1) \textit{Caption Generation} (where we generate captions for the image input associated with the content sample and pass this in the text prompt) and (2) \textit{Direct Image Input} (where we utilize a VLM that can operate on multimodal text + visual input directly). Note that we can also obtain the zero-shot setting as a special case for these multimodal approaches by simply discarding the exemplar set in few-shot learning. 

\subsubsection{Caption Generation (CG)}
We pass the input image(s) (thumbnails) to BLIP pretrained (Bootstrapping Language-Image Pre-training) model on the captioning task~\cite{shen2022how, li2022blip} and extract the generated caption. Then, we augment the textual content instance with this caption. Additionally, we did the same experiment with Qwen-VL-chat model \cite{Qwen-VL}. More details are mentioned in the Appendix~\ref{appendix:qwen_vl}. We essentially follow the same procedure for the text-only FS-ICL approach but with the generated caption also provided as input to the LLM. Essentially, denoting the captioning VLM as ${V}$, the function $f$ incorporates a caption \( c_i \) generated by \( V \) from the visual component, i.e. $c_i = V(v_i)$, as follows:%\vspace{-2mm}
\[
f_{\text{FS-ICL-CG}}(x_i) \coloneq \mathcal{L}(t_i, c_i, \mathcal{E}, \mathcal{P}_{\text{FS-ICL-CG}}).
\]
The goal here is to extract additional signals from the image, which may not be present in the text (for example, a YouTube video may have a harmless caption, but a sexually explicit or clickbait thumbnail). However, errors in the caption generation process might propagate further down the learning pipeline and reduce LLM performance.

\subsubsection{Direct Image Input (DII)}
Some LLMs/VLMs, such as GPT-4o-Mini \cite{OpenAI_GPT-4o_mini} operate on multimodal input directly~\cite{radford2021learning} (i.e., they can process and integrate both visual and textual data simultaneously). Now, by simply utilizing such an VLM in the FS-ICL approach and providing image data along with text input, we can augment task performance. The classification function is as follows:%\vspace{-2mm}
\[
 f_{\text{FS-ICL-DII}}(x_i) \coloneq \mathcal{L}(v_i, t_i, \mathcal{E}, \mathcal{P}_{\text{FS-ICL-DII}}).
\]

The Direct Image Input (DII) approach leverages multimodal data, enhancing the model's understanding by directly incorporating visual cues. This approach can lead to more accurate and context-aware predictions, especially in cases where visual content carries critical information. However, it is relatively computationally intensive.

\section{Experiments and Results}

\subsection{Experimental Setup}

\textbf{Datasets.}
\looseness-1We employ a curated dataset of YouTube videos~\cite{Jo_2024} that encompasses distinct types of harms identifiable in multimodal social media data. Each video in the dataset was labeled as harmful/harmless by crowdworkers, LLMs, and domain experts. The ground truth is taken as the majority label across these three labeling actors.
We used a train-test split of 3,000 videos evenly divided between harmful and harmless videos.
We chose this dataset because of the diverse nature of harms represented as well as features from multiple modalities (text, image).
Additionally, to validate the generalizability of our approach, we evaluate it on two publicly available datasets; the Jigsaw Toxicity Classification Dataset~\cite{jigsaw_unintended_bias_2019}, which consists of comments from the Civil Comments platform, labeled for toxicity and various identity-based biases, and the Measuring Hate Speech dataset~\cite{ucberkeley_dlab_measuring_hate_speech} by UC Berkeley's D-Lab, which contains a large corpus of annotated social media posts specifically labeled for hate speech. More details on all datasets are provided in Appendix\ref{appendix:datasets}.\vspace{1mm}

\noindent\textbf{Models.}
For experiments, we use two open-source LLMs, Mistral-7B and Llama2-13B. Additionally, we employ proprietary closed-source OpenAI LLMs, GPT-3.5-Turbo and GPT-4o-Mini. We use the BLIP~\cite{li2022blip} model for generating captions and LLaVA~\cite{liu2024improved}, GPT-4o-Mini~\cite{OpenAI_GPT-4o_mini} , OpenFlamingo~\cite{Alayrac2022FlamingoAV} for vision-based ICL. All the prompts we design are provided in Appendix \ref{appendix:prompts}.

% \textbf{Implementation.}
% All experiments were conducted on an A6000 GPU environment using available computational resources suitable for running LLMs. Experiments involving LLaVA~\cite{liu2024improved}, OpenFlamingo~\cite{Alayrac2022FlamingoAV} VLMs were conducted on A100 GPU in google colab.

\subsection{Interpretation of Toxicity Scores}
We use toxicity estimates from the Perspective API and the OpenAI moderation API only as auxiliary signals. These services are proprietary black-box systems; their internal models are not publicly documented, but they are widely adopted and trained on large-scale human-annotated data, likely using modern neural architectures. The scores should therefore be interpreted as \emph{perceived likelihood of offensive or harmful language} rather than definitive ground truth. We use the Perspective API \textbf{TOXICITY} attribute rather than SEVERE\_TOXICITY. Prior studies have shown that SEVERE\_TOXICITY exhibits a heavy-tailed distribution with most values close to zero, which can overemphasize a small set of extreme cases and make model comparisons less informative. The TOXICITY attribute provides a broader and better-calibrated measure of harmful language and is the attribute most commonly adopted in moderation benchmarks. This choice aligns the evaluation with general content-moderation objectives rather than only extreme abuse detection. \cite{perspectiveapi_attributes}

In this work, toxicity scores are used solely to (i) stratify evaluation by severity, and (ii) analyze error patterns. They are \textbf{not} used to generate the final moderation decisions of our LLM/VLM systems, nor to train or fine-tune any models. Consequently, our reported performance reflects the behavior of the proposed methods rather than the behavior of these APIs.
\subsection{Baselines}

\looseness-1\textbf{Specific FSL Baselines.}
We considered two state-of-the-art baselines from prior work: \textit{Prototypical Networks}~\cite{snell2017prototypical} and \textit{Matching Networks}~\cite{vinyals2016matching}.
Matching networks involve a similarity-based approach where the aim is to adapt to new classes with minimal labeled data by leveraging the powerful text representations from transformer-based models such as BERT.
Prototypical Networks learn a metric space wherein classification is performed by computing distances to prototype representations of each class.\vspace{1mm}

\noindent\textbf{Proprietary Baselines.}

In addition to the deep learning few-shot baselines, we use two industry-grade moderation modules to compare the performance of our models. Both Perspective API~\cite{perspective_api} and OpenAI Moderation API~\cite{openai_moderation} are prominent tools that use advanced machine learning for content moderation and have widespread adoption in the industry; Perspective is used by platforms like Reddit, the New York Times, and OpenWeb,~\footnote{https://perspectiveapi.com/} while the OpenAI moderation API is used by the OpenAI family of products to detect violating content. Both of these modules produce a score for a given text input, with higher scores reflecting greater perceived harm (e.g., toxicity, insult, threat). 
% %
% The Perspective API~\cite{perspective_api} is a tool developed by Jigsaw to help improve online conversations. It uses machine learning models to identify and measure the level of toxicity in online content. The API is used for content moderation by several leading platforms or websites such as Reddit or the New York Times.
% %
% The OpenAI Moderation~\cite{openai_moderation} API is designed to help developers identify and handle potentially harmful or unsafe content within text.

\subsection{Results} 

\begin{table}[t]
\caption{Performance comparison of various baseline models on various metrics (\%).}
\label{tab:baseline_performance}
\centering
\resizebox{0.48\textwidth}{!}{%
\begin{tabular}{@{}lcccc@{}}
\toprule
\textbf{Model}                   & \textbf{Accuracy} & \textbf{Precision} & \textbf{Recall} & \textbf{F-1} \\ \midrule
Perspective API                  & 50.36             & 50.20              & 98.00           & 66.40        \\
Crowdsourced - Worst             & 53.40             & 53.50              & 53.39           & 53.03        \\
OpenAI Moderation API            & 55.93             & 53.69              & 86.67           & 66.31        \\
Crowdsourced - Majority          & 61.23             & 63.28              & 61.21           & 59.65        \\
Domain Expert                    & 90.97             & 91.93              & 90.96           & 90.91        \\ \bottomrule
\end{tabular}
}
\end{table}

\begin{figure}[t]
\centering
\includegraphics[width=0.49\textwidth]{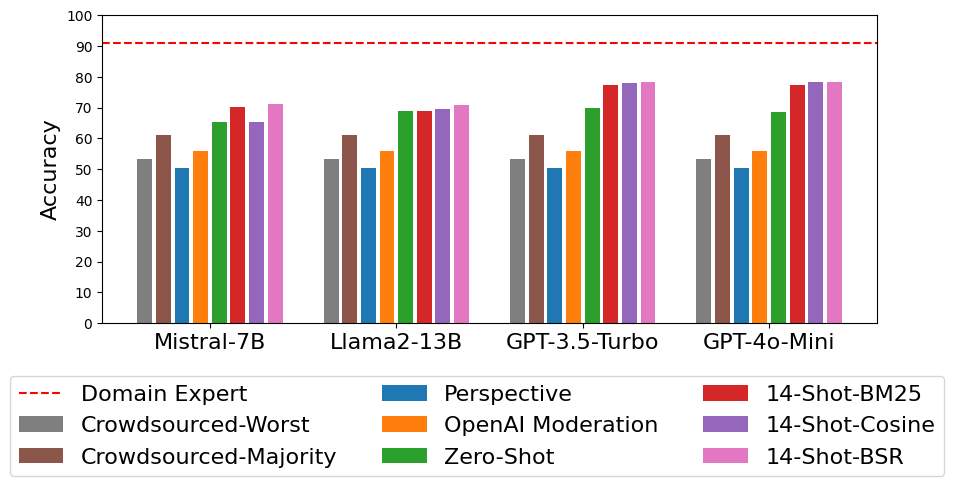}\vspace{-2mm}
\caption{Accuracy (\%) across different models for ZSL and FS-ICL (14-Shot; using BM25, Cosine, BSR selectors), proprietary baselines (Perspective and OpenAI Moderation APIs), crowdsourced annotators (\textit{Crowdsourced-Worst}: minority label; \textit{Crowdsourced-Majority}: majority label), and domain experts.
% Domain Experts achieve the highest accuracy overall. LLMs outperform both proprietary baselines and crowdsourced annotators. FS-ICL (14-Shot) configurations lead to significant accuracy improvements compared to Zero-Shot in models like GPT-3.5-Turbo and GPT-4o-Mini, with GPT-3.5-Turbo achieving the best accuracy of 78.33\% using selector BSR.
}
\label{fig:ICL-bar}\vspace{-4mm}
\end{figure}

\looseness-1\textbf{Zero-Shot Harm Detection.}
% make captions long
In ZSL, we utilize LLMs to classify content as harmful/harmless without using labeled training examples. As shown in Figure~\ref{fig:ICL-bar}, the LLMs significantly outperformed proprietary baselines in terms of accuracy. Specifically, the OpenAI Moderation API achieves 56\% accuracy and Perspective API achieves 50\% accuracy as shown in Table~\ref{tab:baseline_performance}. Our ZSL approach, on the other hand, was able to achieve accuracies of 65\% (Mistral-7B), 69\% (Llama2-13B; GPT-4o-Mini), and 70\% (GPT-3.5-Turbo). We also compare performance across different ZSL settings over various metrics, such as Precision, Recall, and F1 score as shown in Table~\ref{tab:zsl_performance}

\begin{table}[t]
\caption{Performance comparison of models across ZSL, ZSL-CG, and ZSL-DII settings using various metrics (\%).}
\label{tab:zsl_performance}
\centering
\resizebox{0.43\textwidth}{!}{%
\begin{tabular}{@{}lcccc@{}}
\toprule
\textbf{Model}                  & \textbf{Accuracy} & \textbf{Precision} & \textbf{Recall} & \textbf{F-1} \\ \midrule
\multicolumn{5}{c}{\textbf{ZSL}}                                                                          \\ \midrule
Mistral-7B                      & 65.23             & 66.55              & 65.22           & 64.52        \\
Llama2-13B                      & 68.97             & 69.25              & 68.96           & 68.85        \\
GPT-4o-mini                     & 68.67             & 68.75              & 68.67           & 68.63        \\
GPT-3.5-turbo                   & 70.00             & 70.00              & 70.00           & 70.00        \\ \midrule
\multicolumn{5}{c}{\textbf{ZSL CG}}                                                                       \\ \midrule
Mistral-7B                      & 63.41             & 63.58              & 63.60           & 63.41        \\
Llama2-13B                      & 67.93             & 67.81              & 67.65           & 67.68        \\
GPT-4o-mini                     & 68.23             & 69.31              & 69.89           & 69.59        \\
GPT-3.5-turbo                   & 70.20             & 72.96              & 71.34           & 72.14        \\ \midrule
\multicolumn{5}{c}{\textbf{ZSL DII}}                                                                      \\ \midrule
OpenFlamingo                 & 50.00             & 50.00              & 50.00           & 50.00        \\
LLaVa VLM                           & 54.00             & 59.50              & 62.50           & 60.50        \\
GPT-4o-mini          & 70.00             & 70.00              & 72.00           & 71.00        \\ \bottomrule
\end{tabular}
}
\end{table} 

\begin{figure*}[t]
\centering
\includegraphics[width=0.62\textwidth]{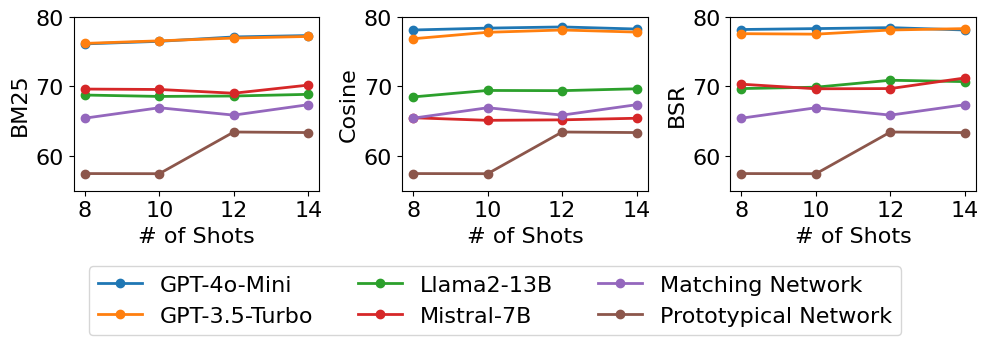}\vspace{-3mm}
\caption{Accuracy (\%) of different LLMs for FS-ICL while varying number of shots and ICL selection methods, and open-source deep learning baselines (Matching Network and Prototypical Network) as well proprietary baselines. Among the selectors, BSR achieves the highest accuracy overall and across LLMs, GPT-4o-Mini achieves the highest performance. Interestingly, the number of shots has only a minor impact on the accuracy.}
\label{fig:ICL-lines}\vspace{-2mm}
\end{figure*}

\begin{figure}[t]
\centering
\includegraphics[width=0.45\textwidth]{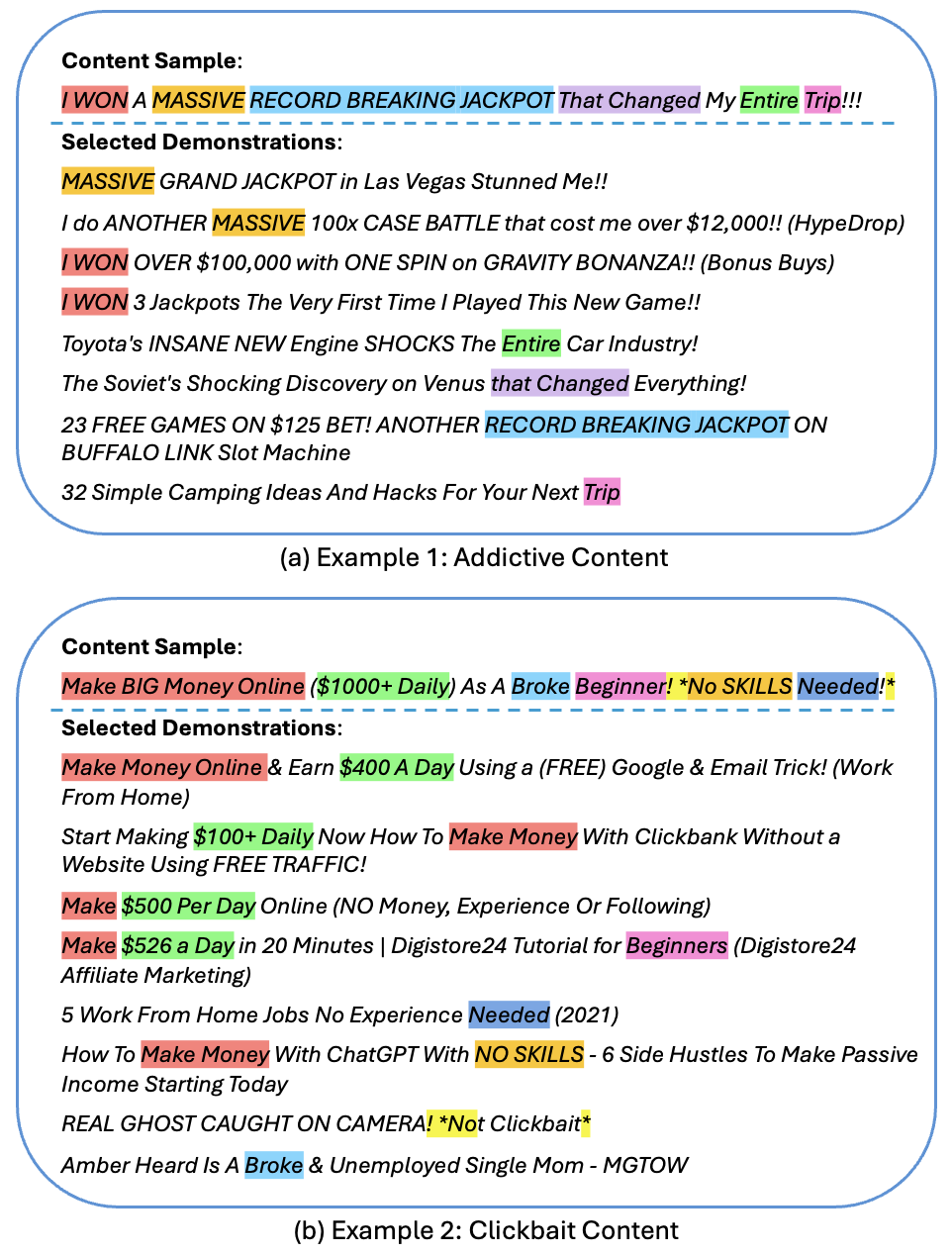}\vspace{-2mm}
\caption{Illustrative examples of few-shot ($k=8$) selection for two harm categories: (a) \textit{addictive gambling} and (b) \textit{financial clickbait}. Colors show how demonstrations cover salient aspects of the samples.}
\label{fig:coverage}\vspace{-5mm}
\end{figure}

\begin{figure}[t]
\centering
\includegraphics[width=0.5\textwidth]{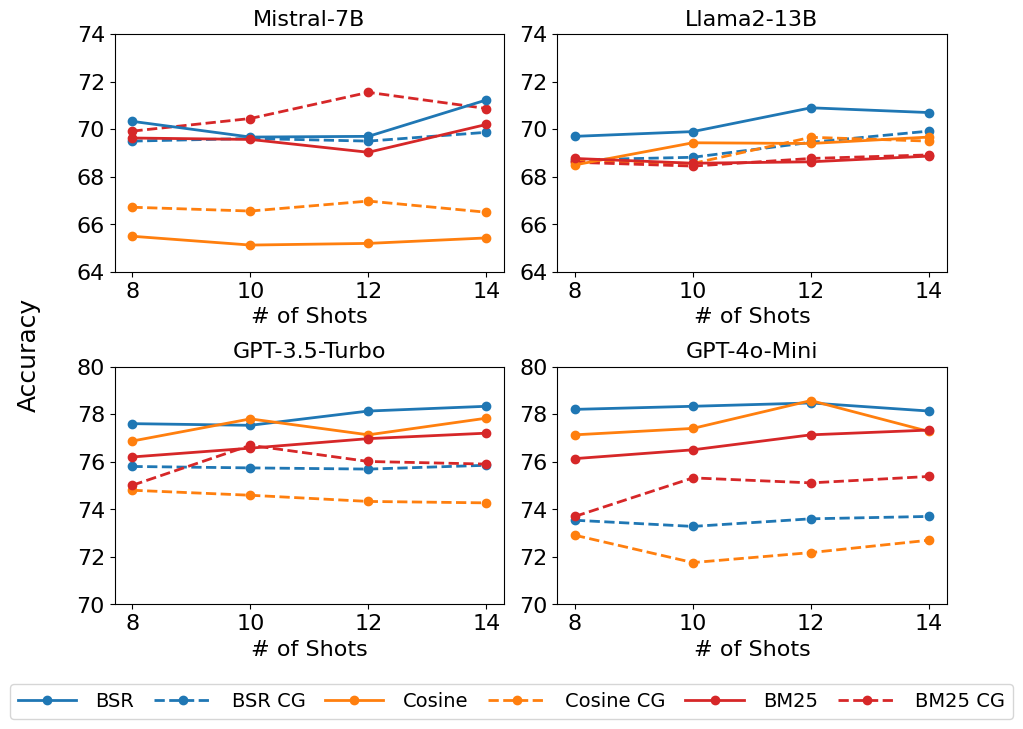}\vspace{-2mm}
\caption{Comparison of FS-ICL and FS-ICL-CG approaches of different models across different shot numbers and selection methods. FS-ICL-CG does not improve upon FS-ICL, indicating that inclusion of captions does not consistently enhance performance.}
\label{fig:cg_lines}\vspace{-4mm}
\end{figure}

\vspace{2mm}%@@@
\noindent\textbf{Few-Shot Harm Detection.} We employ few-shot ICL to leverage the full capacity of LLMs and aim to achieve performance comparable to the domain expert accuracy of 90.67\% (In our study, ground truth labels were set by majority agreement among domain experts, GPT, and crowdworkers. The domain expert accuracy of 90.67\% reflects alignment with this combined majority rather than domain expert labels alone as in \cite{Jo_2024}. The FS-ICL approaches demonstrate a significant improvement over the zero-shot configurations, with a notable 5-10\% increase in accuracy. Among the selectors, BSR consistently achieves the best performance.

LLMs in the few-shot setting also significantly outperform open-source FSL baselines such as Prototypical Networks and Matching Networks. As seen in Figure~\ref{fig:ICL-lines}, GPT-4o-Mini with 12 shots and the BERTScore selector in the FS-ICL approach achieves an accuracy of 78.57\%, the best-performing baseline model, whereas Prototypical Network (roberta-base) using 12 shots, achieves only 67.59\% accuracy. Similarly, GPT-3.5-turbo in the FS-ICL approach achieves an accuracy of 78.33\% using 14 shots and BERTScore, while the Matching Network (bert-base-uncased) with 14 shots achieves a much lower accuracy of 67.38\%. Overall, the closed-source models outperform the open-source LLMs by a significant margin.%\vspace{1mm}

%
%The BERTScore selector consistently enhances model performance across different configurations, proving to be more effective than the distance-based metrics used by Prototypical Networks and Matching Networks and also Cosine and BM25 selectors. 
%
% The FS-ICL-CG approach shows significant improvement over zero-shot results but does not surpass the performance of the FS-ICL approach. 

%\noindent\textbf{Remark.}
We also observe that the number of shots has a minor impact on the accuracy (e.g., 8-shot: 78.2\% vs. 14-shot: 78.13\% for GPT-4o-Mini using BSR). This may be because the number of salient aspects in the content samples, i.e., video titles, was not greater than the number of shots tested, due to YouTube's title length limitations. While 14 shots resulted in slightly better performance, 8 shots was sufficient to cover most of the salient aspects needed for LLMs to determine whether a video is harmful. We offer a few examples to qualitatively demonstrate this in Figure~\ref{fig:coverage}.

\vspace{2mm}\looseness-1\noindent\textbf{Multimodal Harm Detection.} Lastly, we systematically assess whether (and the extent to which) additional improvements can be gained from the integration of both text and visual (i.e. \textit{multimodal}) input. To this end, we utilize our FS-ICL-CG and FS-ICL-DII approaches under various shot configurations. For FS-ICL-CG, we generate captions for the video thumbnail to aid in the classification process. As seen in Figure~\ref{fig:cg_lines}, despite this additional input FS-ICL-CG does not improve upon the performance of FS-ICL and often leads to a reduction in performance (e.g., FS-ICL: 78.13\% vs. FS-ICL-CG: 73.7\%  for GPT-4o-Mini using BSR on a 14-shot setting). This might be due to the generated captions not being able to capture subtle harm information present in the vision domain. Additional results, including other metrics, are provided in Table~\ref{tab:fsl_cg_best_performers}.

\begin{table}[t]
\caption{FSL-CG performance comparison for models using various metrics (\%).}
\label{tab:fsl_cg_best_performers}
\centering
\resizebox{0.46\textwidth}{!}{%
\begin{tabular}{@{}llrrrr@{}}
\toprule
\textbf{Model}                                          & \textbf{Accuracy} & \textbf{Precision} & \textbf{Recall} & \textbf{F-1} \\ \midrule
{GPT-3.5-turbo}                           & 76.69             & 77.39              & 77.14           & 76.67        \\
{GPT-4o-mini}                           & 75.38             & 76.21              & 75.87           & 75.35        \\
{Mistral-7B}       & 71.55             & 73.08              & 70.62           & 70.41        \\
{Llama2-13B}               & 69.92             & 72.41              & 70.82           & 69.58        \\ \bottomrule
\end{tabular}%
}
\end{table}
%Note that FS-ICL-CG approaches improves upon the zero-shot configurations, with a notable 5-10\% increase in accuracy.
%

% all experiments 

%\subsubsection{FS-ICL-DII}
%To address the limitations of FS-ICL-CG approach and fully leverage the capability of LLMs to incorporate multimodal for harm detection, we experiment on the FS-ICL-DII approach, which allows the integration of both text and direct image input. 
%Lastly, to test if additional improvements to the accuracy can be gained from the integration of both text and direct image input, we experiment on the FS-ICL-DII approach. In this way, we fully leverage the capability of LLMs to incorporate multimodal message features for harm detection. 
\looseness-1Next, we utilize the FS-ICL-DII approach where we provide visual input directly to natively multimodal LLMs. We consider the OpenFlamingo, LLaVA, GPT-4o-Mini LLMs. The results for these are shown in Table~\ref{tab:FSICLDII}. We see that GPT-4o-Mini in the 14-shot configuration achieves the best performance, outperforming all other models across various settings and attaining an accuracy of 80\%.
For GPT-4o-Mini specifically, it is clear that the LLM benefits greatly from the additional multimodal examples provided. However, this is not the case for LLaVa and OpenFlamingo. Performance is fairly low throughout ($\approx 50\%$), even after few-shot ICL in OpenFlamingo (LLaVa does not directly support ICL so results are omitted). Hence, the choice of the LLM is very important for multimodal learning, especially in the context of harm classification. We provide additional results comparing performance using various metrics as shown in Table~\ref{tab:fsl_dii_best_performers}

The performance of various models across different configurations is summarized in the Appendix~\ref{appendix:qwen_vl},~\ref{appendix:other_metrics} and tables~\ref{tab:qwen_vl_captioning_results}~\ref{tab:fsl_title_performance}, showcasing their effectiveness on the FSL tasks. These results highlight key metrics such as accuracy, precision, recall, and F-1 scores, providing a comparative analysis of the models. For a detailed breakdown and additional results, refer to the Appendix~\ref{appendix:qwen_vl},~\ref{appendix:other_metrics}, where we present extended tables on model configurations and performance trends.

\begin{table}[t]
\caption{Accuracy (\%) of LLMs for varying the number of shots for the {FS-ICL-DII} approaches. N/A denotes lack of native support for few-shot learning.}%\vspace{-3mm}
\label{tab:FSICLDII}
\centering
\resizebox{0.38\textwidth}{!}{%
\begin{tabular}{@{}cccccc@{}}
\toprule
\multirow{2}{*}{\textbf{Model}} & \multicolumn{5}{c}{\textbf{Shots}} \\ \cmidrule(l){2-6} 
                                & 0    & 8     & 10    & 12   & 14   \\ \midrule
{OpenFlamingo}           & 50   & 50    & 52    & 54   & 54   \\
{LLaVa VLM}              & 54   & N/A   & N/A   & N/A  & N/A  \\
{GPT-4o-mini}            & \textbf{70}   &\textbf{79}    & \textbf{79}    & \textbf{79}   & \textbf{80}   \\ \bottomrule
\end{tabular}
}
\end{table}

\begin{table}[t]
\caption{FSL-DII Best Performers on multimodal classification using various metrics (\%).}
\label{tab:fsl_dii_best_performers}
\centering
\resizebox{0.44\textwidth}{!}{%
\begin{tabular}{@{}ccccc@{}}
\toprule
\textbf{Model}        & \textbf{Accuracy} & \textbf{Precision} & \textbf{Recall} & \textbf{F-1} \\ \midrule
{OpenFlamingo} & 54.0                & 62.5               & 69.0              & 65.5         \\
{LLaVa VLM}    & 54.0                & 59.5               & 62.5            & 60.5         \\
{GPT-4o-mini}  & 80.0                & 82.0                 & 80.0              & 81.0           \\ \bottomrule
\end{tabular}
}
\end{table}

\subsection{Statistical Analysis}
\label{sec:stat_analysis}

We assess statistical reliability in two complementary ways.
\newline
\textbf{(1) Distributional difference tests:} We apply McNemar's paired test to compare prediction behaviors across configurations Appendix~\ref{appendix:stat_analysis}. These tests assess whether two systems make significantly different decisions on the same samples, but they do \emph{not} quantify which system is more accurate. 
\newline
\textbf{(2) Performance effect estimation:} To support claims of higher accuracy and F1, we estimate 95\% confidence intervals via bootstrap resampling (1,000 iterations) over the test set. The deltas in Table~\ref{tab:effect_sizes_full} are computed from the same fixed test splits used in Tables 4, 13, and 17, with deterministic decoding (temperature=0) and no re-tuning. This two-stage analysis separates behavioral divergence from performance gains and reduces over-interpretation of small numeric differences. 

\paragraph{Interpretation of observed effects.}
Applying these criteria to Tables~\ref{tab:effect_sizes_full} and~\ref{tab:dataset_effects}, we observe that FS-ICL yields consistently large improvements over ZSL across datasets, while model-to-model differences under identical FS-ICL settings remain minimal (Table~\ref{tab:model_diffs}). These results indicate that the dominant factor is the few-shot paradigm and selector strategy rather than the specific LLM family.

\section{Discussion}

\looseness-1\noindent\textbf{LLM Performance Across Other Harm Classification Datasets.} To ensure that LLM performance at harm classification is not localized to the YouTube videos dataset \cite{Jo_2024}, we experiment on two additional harm-specific datasets: D-Lab Hate Speech \cite{ucberkeley_dlab_measuring_hate_speech} and Jigsaw Toxicity \cite{jigsaw_unintended_bias_2019}. These datasets do not consider diverse harm categories as the YouTube dataset (i.e. focus solely on hate speech and toxicity). As the GPT-4o-Mini and GPT-3.5-Turbo models were the highest performers in prior experiments, we use these to undertake experiments on the additional datasets. These results are shown in Appendices \ref{appendix:toxic} and \ref{appendix:hate}. We find that the LLMs outperform the other baselines on these benchmarks as well, highlighting their use as content moderators. 

\begin{figure}[t]
\centering
\includegraphics[width=0.48\textwidth]{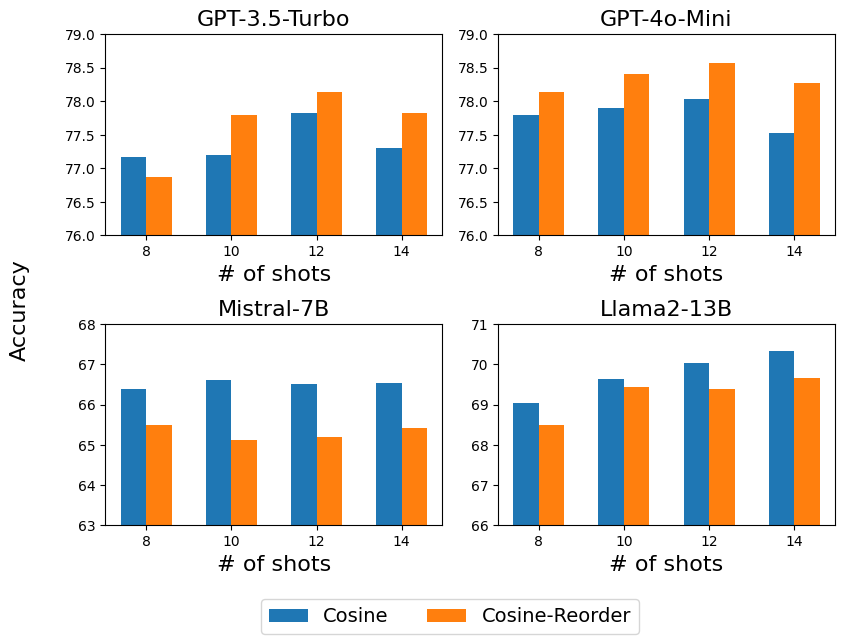}\vspace{-3mm}
\caption{Impact of reordering on accuracy (\%).}
\label{fig:reorder}%\vspace{-3mm}
\end{figure}

\begin{figure}[t]
\centering
\includegraphics[width=0.48\textwidth]{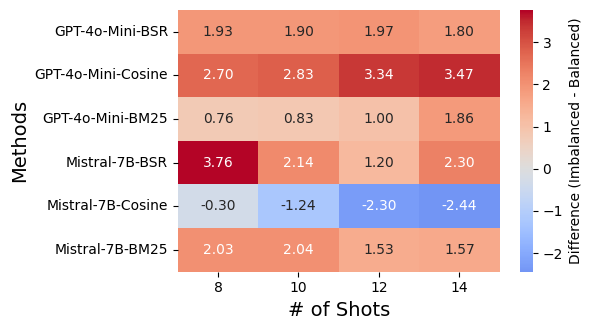}\vspace{-3mm}
\caption{Accuracy difference (\%) between \textit{balanced} and \textit{imbalanced} demonstration selection for GPT-4o-Mini and Mistral-7B across varying shots and selection methods. Red denotes \textit{imbalanced} $>$ \textit{balanced}, while blue denotes the converse.}\vspace{-6mm}
\label{fig:balance}
\end{figure}

\vspace{2mm}\noindent\textbf{Exemplar Reordering and Balanced Selection.} Since ICL and few-shot learning lead to greatly improved performance in harm classification, we undertake further ablations to analyze how exemplar order and balanced selection affect model performance. As exemplar order has been shown to influence performance \cite{lu2021fantastically}, we first undertake two experiments for exemplar selection (using cosine similarity as the selector): (1) reordering exemplars based on their instance-level metric, which means the most similar demonstration lists closest to the content sample, and (2) listing exemplars in the prompt as selected by the coverage-based metric, without reordering. Figure \ref{fig:reorder} details the results. As seen, the GPT models benefit from reordering but this is not the case for Mistral and Llama, which prefer the original ordering based on the selection metric. Thus, it is important to reorder depending on the model used. Next, we conduct two experiments with exemplar selection: (1) explicitly balancing the number of examples from each class (\textit{balanced} selection), and (2) selecting examples based solely on a coverage-based selector metric, regardless of class distribution (\textit{imbalanced} selection) and using the Mistral-7B and GPT-4o-Mini models, to show the impact of class-balanced selection. Figure \ref{fig:balance} shows that imbalanced selection generally results in better performance across the board.

\vspace{2mm}\looseness-1\noindent\textbf{Utilizing More Descriptive Prompts.} We also compare the performance effect of using more descriptive prompts. Currently, our prompts do not include specific definitions of harm categories~\cite{Jo_2024}, so we include these and run experiments on the YouTube dataset with GPT-4o-Mini. The detailed prompts and results are provided in Appendix \ref{appendix:descriptive}. We find the more detailed prompts do not impact LLM performance much and the best gain is only 0.37\%. Hence, it might be better to use more concise prompts and under-utilize the context, thereby enabling the model to understand the definition of harm via FS-ICL.

%@@@
% While our study focuses on open-source LLMs and VLMs, we acknowledge that they often underperform compared to proprietary models like GPT-4 or Gemini. This gap arises from factors such as limited fine-tuning on moderation-specific tasks, smaller and less diverse training datasets, and the absence of advanced optimization techniques available to commercial providers. Proprietary models also benefit from continuous updates and reinforcement learning with human feedback (RLHF), enhancing their robustness. Our work focuses solely on prompting-based approaches and does not explore fine-tuning or RLHF strategies for improving model performance. This choice was intentional, as fine-tuning and RL require substantial computational resources, time, and access to large task-specific datasets, which are beyond the scope of this study. While LLMs are often seen as less scalable compared to smaller models like BERT, the few-shot prompting approach we employ reduces the need for costly retraining, helping mitigate scalability concerns.

While our study focuses on open-source LLMs and VLMs, we acknowledge that they often underperform compared to proprietary models such as GPT-4 or Gemini. This gap likely stems from several factors: (i) proprietary systems are trained on substantially larger and more diverse datasets, including platform-specific safety data that is not publicly available; (ii) they benefit from extensive instruction tuning and reinforcement learning with human feedback (RLHF) aligned with moderation objectives; and (iii) multimodal safety requires high-quality cross-modal grounding and curated image–text harm datasets, which remain limited in the open-source ecosystem. In addition, commercial deployments typically integrate ensemble moderation signals, dynamic guardrails, and continual updates, whereas stand-alone open models rely on static checkpoints.

Our comparison targets widely deployed industrial baselines rather than the full landscape of fine-tuned classifiers. This reflects the scenario where retraining pipelines are unavailable and practitioners rely on off-the-shelf systems. Fine-tuned models remain complementary when curated data and compute are accessible.

Our work intentionally focuses on prompting-based approaches without fine-tuning or RLHF, as these require substantial computational resources and access to large task-specific datasets. Nevertheless, the results suggest several concrete paths to narrow the gap: releasing larger open multimodal safety datasets, developing instruction-tuning recipes targeted at harm detection, employing retrieval-augmented in-context learning to inject policy knowledge, and hybrid pipelines that combine open LLM reasoning with external safety classifiers. The strong gains from FS-ICL-DII indicate that improved demonstration selection and structured prompting can already mitigate part of the disparity without retraining.

\section{Conclusion}
% % Write conclusion after everything else is done
% Our experiments underscore the value of few-shot learning in social media harm detection by enhancing the performance of both LLMs and VLMs, especially when combined with multimodal inputs. While zero-shot configurations demonstrate strong baseline capabilities, incorporating a small number of labeled examples significantly improves performance, enabling models to generalize better across diverse types of content. The effectiveness of multimodal approaches in the FS-ICL-DII configuration particularly stands out, showing that advanced models like GPT-4o-Mini can leverage both visual and textual data to achieve superior performance in harm detection tasks. In future research, we plan to further explore optimizing caption generation, the use of more context in case of Youtube videos (e.g., providing more frames) and exploring other sophisticated multimodal models to maximize the potential, trying other approaches and subsclasses of harm detection.

We study online harm classification (or content moderation) problem where approaches need to be highly effective while requiring minimal supervision (i.e. annotated data) and should adapt to the ever-changing dynamic nature of harmful content. We employ LLMs for this task and demonstrate that LLM performance in the few-shot in-context learning setting (i.e. minimally supervised + dynamic) can near domain expert performance. We utilize open-source and relatively inexpensive closed-source LLMs across different size scales (Mistral-7B, Llama2-13B, GPT-3.5-Turbo, GPT-4o-Mini) to ensure that the approach can scale to actual moderation of social media content. Further, we analyze LLMs in the multimodal setting and find that performance can be greatly improved, but only for closed-source multimodal LLMs such as GPT-4o-Mini and not smaller open-source variants (LLaVa and OpenFlamingo). Our findings underscore the the benefits of LLMs as a better alternative to proprietary baselines (Perspective and OpenAI Moderation APIs) and other deep learning few-shot baselines (e.g. Prototypical Networks) for harm identification.

\section{Limitations}
We highlight limitations across generalizability, prompt uncertainty, dataset bias, evaluation scope, and deployment constraints. While our results demonstrate the promise of LLM- and VLM-based moderation, generalizability is constrained because our experiments focus primarily on English-language datasets and a limited set of social media domains; harmful expression varies across languages, cultures, and platforms, and performance may not transfer to low-resource languages, regional slang, or communities with different norms. In addition, the use of static benchmarks leaves open questions about robustness to distribution shift, emerging harm types, and adversarial manipulation in real-world streams. Model behavior is also sensitive to prompt wording, demonstration selection, and input formatting, introducing prompt uncertainty that can affect reproducibility and stability of decisions. The datasets used contain annotator subjectivity and historical biases, and reducing multi-label signals to binary decisions can obscure nuanced cases, limiting the interpretability of measured improvements. Moreover, toxicity scores originate from proprietary models with unknown biases and may conflate stylistic aggression with genuine harm; analyses based on them should therefore be interpreted cautiously. Our evaluation emphasizes accuracy and F1 on test sets and does not consider user-level impacts, cost-sensitive trade-offs, or long-tail harms that are critical in deployment. Furthermore, multimodal inference in FS-ICL-DII is computationally expensive (typically 10–30 seconds per sample), posing scalability challenges for real-time moderation. Finally, automated moderation lacks guaranteed interpretability and may risk unjustified censorship or under-detection without human oversight, transparent auditing, and appeal mechanisms. The current performance gap between open- and closed-source multimodal LLMs highlights the need for more transparent training data, safety-focused alignment methods, and community benchmarks tailored to harm detection. Future work should examine multilingual settings, prompt robustness, cost-aware metrics, and human–AI collaboration for safer and more reliable moderation.

\section{Ethics Statement}
%@@@
% Our work explores the application of LLMs for detecting harmful content in YouTube videos, aiming to support the development of safer online communities. We conducted experiments using publicly available datasets and ensured that no private or personally identifiable information was used. By sharing our code and implementation details in the Appendix  \ref{appendix:code}, we promote transparency and reproducibility. We recognize the importance of ongoing research into the ethical aspects of automated content moderation, including fairness, accuracy, and user impact, and hope our contributions encourage further advancements in responsible and effective harm detection methods.
This study uses a public YouTube dataset containing only video titles and thumbnails, the Berkeley D-Lab Measuring Hate Speech corpus of Twitter posts, and the Jigsaw Toxicity dataset of Wikipedia comments. No private, non-public, or personally identifiable information was accessed, and we follow the dataset’s terms of use. We do not attempt to identify creators or users. Harm labeling is subjective and culturally dependent. The ground-truth labels were produced by crowdworkers, domain experts, and LLMs, which may reflect biases and disagreement. Our models can inherit these biases and may misclassify content from marginalized communities or non-Western contexts. Automated moderation carries risks. False positives can lead to unjust censorship and reduced visibility of legitimate speech, while false negatives may expose users to harmful material. We do not recommend fully automated enforcement without human review, transparency, and appeal mechanisms. Future work should study fairness across languages and cultures and evaluate real-world impacts before deployment.

\section*{Acknowledgments}
This project was funded with support from the Plurality Institute and The Benjamin Franklin Society Library grant \#88801199.

\bibliography{refs}
% \bibliography{aaai25}

%\clearpage
%\input{Checklist}
\appendix

\section*{Appendix}

\section{Datasets}\label{appendix:datasets}

\subsection{YouTube Harms Dataset}
We leverage a multimodal dataset~\cite{Jo_2024}, comprising of YouTube videos from a variety of harm categories. The data is categorized online harm into six types: information harms, hate and harassment harms, clickbaitive harms, addictive harms, sexual harms, and physical harms. Videos were labeled by crowdworkers (from Amazon MTurk), fine-tuned LLMs and domain experts trained in the social sciences, specifically with a focus on communication and digital media. Each video was classified harmful or harmless, and assigned one or more harm categories if classified as harmful. The final dataset was formed by filtering for the videos where there was agreement between the crowdworker and domain expert, resulting in ~$19000$ labeled videos.
\begin{table}[h!]
\centering
\caption{Distribution of Harm Categories}
\begin{tabular}{ll}
\toprule
\textbf{Harm Category} & \textbf{Percentage} \\
\midrule
Information harms           & 15.2\% \\
Hate and harassment harms   & 12.6\% \\
Clickbait harms             & 17.7\% \\
Addictive harms             & 15.5\% \\
Sexual harms                & 8.5\% \\
Physical harms              & 15.6\% \\
Harmless                    & 14.9\% \\
\bottomrule
\end{tabular}
\end{table}

\subsection{Measuring Hate Speech Corpus (D-Lab)}
The Berkeley D-Lab Measuring Hate Speech Corpus~\cite{ucberkeley_dlab_measuring_hate_speech} is a labeled dataset curated to facilitate the study and measurement of hate speech in online platforms. The dataset is composed of $6,954$ social media posts, which have been manually annotated for the presence of hate speech. The corpus is derived from Twitter data collected from a period between January 2015 and June 2018. Each post has been labeled according to three key categories: (i) hate speech, (ii) offensive but not hate speech, and (iii) neither offensive nor hate speech. The labeling process was conducted by multiple annotators, ensuring inter-annotator agreement and consistent classification standards. The dataset aims to capture a wide range of hate speech expressions, including slurs, dehumanizing language, and threats, across various contexts. Each tweet was annotated by multiple annotators.

\subsection{Jigsaw Toxicity Dataset}

The Jigsaw Toxicity Classification Dataset~\cite{jigsaw_unintended_bias_2019} was developed as part of a Kaggle competition in 2017, with the goal of developing and evaluating machine learning models to detect toxic comments online. This dataset contains a large collection of Wikipedia comments, each labeled by human raters for various types of toxic behavior (toxic, severe toxic, obscene, threat, insult, identity hate). In our experiments, we retrieve the comment text by the comment ID and convert the probability scores for the various toxicity types into a single binary label: Harmful or Harmless.

\section{Selectors}
\label{appendix:selectors}

\subsection{Cosine}
\label{appendix:cosine}
Cosine similarity is a similarity measure between two embedding vectors. We utilize the \textit{Sentence Transformers (SBERT)} library \cite{reimers2019sentence} with the all-mpnet-base-v2 model from Hugging Face to generate sentence embeddings. The cosine similarity between each demonstration and the content sample is then calculated as the dot product of their embeddings, divided by the product of their magnitudes:
\begin{center}
    $
    \text{cosine similarity} = \frac{Q \cdot D}{\|Q\| \|D\|}
    $
\end{center}
Here, $Q$ is the embedding of the content sample and $D$ is the embedding of the demonstration.

\subsection{BM25}
\label{appendix:bm25}
BM25 (Best Matching 25) is a ranking function commonly used in information retrieval systems to rank documents based on their relevance to a given query. Developed as a part of the Okapi system, BM25 considered term frequency saturation and document length normalization, has been tested as an effective information retrieval method. \cite{robertson1995okapi,jones2000probabilistic}
We use the BM25Okapi method from the \textit{rank\_bm25} library with the syntactic structure of size-4 n-grams. The BM25 score between each demonstration and the content sample is calculated as follows:
\begin{center}
\small
    $
    \text{BM25}(Q, D) = \sum_{i=1}^{n} \text{IDF}(q_i) \cdot \frac{f(q_i, D) \cdot (k_1 + 1)}{f(q_i, D) + k_1 \cdot (1 - b + b \cdot \frac{|D|}{\text{avgdl}})}
    $
\end{center}
Here, $Q$ is the content sample and $D$ is the demonstration. $f(q_i, D)$ is the frequency of the n-gram structural element $q_i$ of the content sample in the demonstration candidates. $|D|$ is the length of the demonstration. $k_1$ and $b$ are hyperparameters of the BM25 Okapi class that are set to 1.5 and 0.75 respectively. The $\text{avgdl}$ parameter is the average demonstration length, and $\text{IDF}$ is the weight of the structural element $q_i$.

\subsection{BERTScore}
\label{appendix:bertscore}
BERTScore measures the similarity between two sentences by computing the similarity of each token of the candidate sentence and each token of the reference sentence by cosine similarity. Unlike cosine similarity which compares two sentences directly by the embedding vectors, BERTScore brings the comparison process to the token level. We use bert\_score library \cite{zhang2019bertscore} with the deberta-large-mnli model\cite{he2020deberta} to generate token embeddings. In our experiments, we use only recall as the BERTScore metric and do not apply IDF weighting. We calculate BERTScore-Recall (BSR) between a content sample and a demonstration by first computing the cosine similarity between the token embeddings of both and then maximizing the aggregated cosine similarity score. This can be represented as:
\begin{center}
    $
    \text{BSR}(Q, D) =\sum_{i=1}^{|Q|} \max_{j} \text{cosine similarity}({Q}_i,{D}_j)
    $
\end{center}
Here, $|Q|$ is the length of the content sample, $Q_i$ is the embedding of the $i$-th token of the content sample, and $D_j$ is the embedding of the $j$-th token of the demonstration.

\section{Prompts}
\label{appendix:prompts}

\subsection{ZSL Prompt}
\label{appendix:zslprompt}
% models | prompt
% 1. templates
% 2. exmpales
In the ZSL approach, each prompt consists of the task instruction for harm detection and a single content sample to be classified. The format of these prompts varies depending on the model used. In the absence of relevant examples that offer context or a desired output format, a more restrictive prompt is employed to guide the output.

\subsubsection{Mistral-7B}
\textit{Your task is to classify YouTube videos as Harmful or Harmless based on their metadata. \\
Title: \{title\} \\
Classification:}
\subsubsection{Llama2-13B }
\textit{\textless{}s\textgreater{}[INST] \textless{}\textless{}SYS\textgreater{}\textgreater
        Your task is to evaluate YouTube videos as Harmful or Harmless based on their metadata. You must respond with only one word: "Harmful" or "Harmless," without any additional commentary or explanation.
        \textless{}\textless{}/SYS\textgreater{}\textgreater{} \\
        Title: \{title\} \\
        Classification: [/INST]}
        
\subsubsection{GPT-3.5-Turbo}
{\raggedright
\textit{\{'role': 'user','content': Your task is to classify YouTube videos as Harmful or Harmless based on their metadata. You must respond with only one word: "Harmful" or "Harmless" without any additional commentary or explanation.'\},\\
\{'role': 'user', 'content': 'Title: \{title\}\textbackslash nClassification:'\}}
}
\subsubsection{GPT-4o-Mini}

\textit{\{'role': 'user','content': Your task is to classify YouTube videos as Harmful or Harmless based on their metadata. You must respond with only one word: "Harmful" or "Harmless" without any additional commentary or explanation.'\},\\
{\raggedright\{'role': 'user', 'content': 'Title: \{title\}\textbackslash nClassification:'\}}}

\subsubsection{LLaVa}
\textit{\textbackslash "USER:\textless image\textgreater\textbackslash\textbackslash nAnalyze the given image carefully. Identify any potential risks, safety concerns, sexual explicit content, signs of violence or aggression, indications of illegal activities, health hazards, environmental hazards, or presence of weapons or dangerous objects based on visible details. Consider factors such as the context, setting, activities, expressions, and objects present in the image. Analyze and assess the environment and any notable elements. Identify any potential safety hazards or risks present. Note the presence and actions of people, if any. Check if there is any kind of violence. Finally, classify the overall image context as either 'harmful' if there is any indication of danger, or 'harmless' if it appears safe with no visible threats. Provide your classification as either 'harmful' or 'harmless' and do not give any extra text but the classification.\textbackslash\textbackslash nASSISTANT:\textbackslash "\textbackslash n"}

\subsubsection{OpenFlamingo}
{\raggedright
\textit{f\textbackslash"\textless image\textgreater An image of a close up of two women wrestling in a cage. Classification:harmful\textless $|$endofchunk$|$\textgreater \textless image\textgreater An image of several people standing around a patio with a table and chairs. Classification:harmless\textless $|$endofchunk$|$\textgreater \textless image\textgreater An image of a close up of a woman in a short skirt and heels. Classification:harmful\textless $|$endofchunk$|$\textgreater \textless image\textgreater An image of a man in a headset is talking into a microphone. Classification:harmless\textless $|$endofchunk$|$\textgreater \textless image\textgreater An image of a person looking at a city with a giant explosion. Classification:harmful\textless $|$endofchunk$|$\textgreater \textless image\textgreater An image of three women sitting at a table with a variety of food. Classification:harmless\textless $|$endofchunk$|$\textgreater \textless image\textgreater An image of a man riding a bike on a block of concrete. Classification:harmful\textless $|$endofchunk$|$\textgreater \textless image\textgreater An image of a man in a blue shirt and jeans standing in a parking lot. Classification:harmless\textless $|$endofchunk$|$\textgreater \textless image\textgreater \allowbreak{}\{caption\}. Classification:\textbackslash"}
}
\subsubsection{GPT-4o-Mini (Multimodal)}
\textit{\{"role": "system", "content": "You are a helpful image classification assistant. You must classify given image analyzing the image and Title whether it is harmful or harmless."\}}\\
\textit{\{"role": "user", "content":   {"type": "text", "text": "You are a helpful image classification assistant. You must classify given image whether it is harmful or harmless analyzing the image and Title:{Title}. Harmful: This category includes images depicting sexuality, gambling, violence, weapons, click bait, spam, harmful, or inappropriate content. Examples might be scenes of physical violence, display of sexual affection, weapons, sharp objects, explicit content, spam content, gambling, click bait content or dangerous situations. Harmless: This category includes images that are non-violent, safe, and appropriate. Examples might be everyday scenes, landscapes, or social gatherings without harmful context. If you are not sure, pick whatever label you think is more probable based on your analysis. Give the output in one word if is is harmful or harmless. Do not explain anything."}\}}

% \begin{table*}[ht]
%     \centering
%     \small % Adjust the font size for a better fit
%     \begin{tabularx}{\textwidth}{@{}p{2.5cm}X@{}}
%         \toprule
%         \textbf{Model} & \textbf{Prompt} \\
%         \midrule
%         \textbf{Mistral-7B} & 
%         \texttt{Your task is to classify YouTube videos as Harmful or Harmless based on their metadata.}\\
%         & \texttt{Title: \{title\} }\\
%         & \texttt{Classification:}  \\
%         \midrule
%         \textbf{Llama2-13B} & 
%         \texttt{<s>[INST] <<SYS>>
%         Your task is to classify YouTube videos as Harmful or Harmless based on their metadata.
%         <</SYS>>}\\
%         & \texttt{Title: \{title\}} \\
%         & \texttt{Classification: [/INST]} \\
%         \midrule
%         \textbf{GPT-3.5-Turbo} & 
%         \texttt{\{'role': 'user','content': 'Your task is to classify YouTube videos as Harmful or Harmless based on their metadata.'\},} \\
%         & \texttt{\{'role': 'user', 'content': 'Title: \{title\}\textbackslash nClassification:'\} }\\
%         \midrule
%         \textbf{GPT-4o-Mini} & 
%         \texttt{\{'role': 'user','content': 'Your task is to classify YouTube videos as Harmful or Harmless based on their metadata.'\},} \\
%         & \texttt{\{'role': 'user', 'content': 'Title: \{title\}\textbackslash nClassification:'\} }\\
%         \bottomrule
%     \end{tabularx}
%     \caption{Zero-Shot Prompt Templates used for each model in harm classification.}
%     \label{tab:zsl-prompts}
% \end{table*}

\subsection{FS-ICL Prompt}
\label{appendix:fslprompt}
% models | selector | prompt
% 1. templates
% 2. exmpales
As shown in Table~\ref{tab:fsicl-prompts}, the FS-ICL prompts include a few examples in addition to the ZSL template components. The examples consist of video metadata (title) paired with their corresponding classification.

\begin{table*}[p]
    \caption{FS-ICL Prompt Templates used for each model in harm detection (4-shot example).}
    \label{tab:fsicl-prompts}
    \centering
    \small % Adjust the font size for a better fit
    \begin{tabularx}{\textwidth}{@{}p{2.5cm}X@{}}
        \toprule
        \textbf{Model} & \textbf{Prompt} \\
        \midrule
        \textbf{Mistral-7B} & 
        \texttt{Your task is to classify YouTube videos as Harmful or Harmless based on their metadata.}\\
        & \texttt{Title: \{title\} }\\
        & \texttt{Classification: \{classification\}}  \\
        & \texttt{Title: \{title\} }\\
        & \texttt{Classification: \{classification\}}  \\
        & \texttt{Title: \{title\} }\\
        & \texttt{Classification: \{classification\}}  \\
        & \texttt{Title: \{title\} }\\
        & \texttt{Classification: \{classification\}}  \\
        & \texttt{Title: \{title\} }\\
        & \texttt{Classification: }  \\
        \midrule
        \textbf{Llama2-13B} & 
        \texttt{<s>[INST] <<SYS>>
        Your task is to classify YouTube videos as Harmful or Harmless based on their metadata. You must respond with only one word: "Harmful" or "Harmless" without any additional commentary or explanation.
        <</SYS>>}\\
        & \texttt{Title: \{title\}} \\
        & \texttt{Classification: \{classification\}[/INST]</s><s>[INST] Title: \{title\}} \\
        & \texttt{Classification: \{classification\}[/INST]</s><s>[INST]  Title: \{title\}} \\
        & \texttt{Classification: \{classification\}[/INST]</s><s>[INST]  Title: \{title\}} \\
        & \texttt{Classification: \{classification\}[/INST]</s><s>[INST]  Title: \{title\}} \\
        & \texttt{Classification: [/INST]}\\
        \midrule
        \textbf{GPT-3.5-Turbo} & 
        \texttt{\{'role': 'user','content': 'Your task is to classify YouTube videos as Harmful or Harmless based on their metadata.'\},} \\
        & \texttt{\{'role': 'user', 'content': 'Title: \{title\}\textbackslash nClassification:'\} }\\
        & \texttt{\{'role': 'assistant', 'content': \{classification\}\}}\\
        & \texttt{\{'role': 'user', 'content': 'Title: \{title\}\textbackslash nClassification:'\} }\\
        & \texttt{\{'role': 'assistant', 'content': \{classification\}\}}\\
        & \texttt{\{'role': 'user', 'content': 'Title: \{title\}\textbackslash nClassification:'\} }\\
        & \texttt{\{'role': 'assistant', 'content': \{classification\}\}}\\
        & \texttt{\{'role': 'user', 'content': 'Title: \{title\}\textbackslash nClassification:'\} }\\
        & \texttt{\{'role': 'assistant', 'content': \{classification\}\}}\\
        & \texttt{\{'role': 'user', 'content': 'Title: \{title\}\textbackslash nClassification:'\} }\\
        \midrule
        \textbf{GPT-4o-Mini} & 
        \texttt{\{'role': 'user','content': 'Your task is to classify YouTube videos as Harmful or Harmless based on their metadata.'\},} \\
        & \texttt{\{'role': 'user', 'content': 'Title: \{title\}\textbackslash nClassification:'\} }\\
        & \texttt{\{'role': 'assistant', 'content': \{classification\}\}}\\
        & \texttt{\{'role': 'user', 'content': 'Title: \{title\}\textbackslash nClassification:'\} }\\
        & \texttt{\{'role': 'assistant', 'content': \{classification\}\}}\\
        & \texttt{\{'role': 'user', 'content': 'Title: \{title\}\textbackslash nClassification:'\} }\\
        & \texttt{\{'role': 'assistant', 'content': \{classification\}\}}\\
        & \texttt{\{'role': 'user', 'content': 'Title: \{title\}\textbackslash nClassification:'\} }\\
        & \texttt{\{'role': 'assistant', 'content': \{classification\}\}}\\
        & \texttt{\{'role': 'user', 'content': 'Title: \{title\}\textbackslash nClassification:'\} }\\
        \bottomrule
    \end{tabularx}
\end{table*}

\subsection{FS-ICL-CG Prompt}
\label{appendix:fslprompt}
% models | selector | prompt
% 1. templates
% 2. exmpales
Based on the FS-ICL approach, the FS-ICL-CG prompt integrates image captions associated with the examples and the content sample, as illustrated in Table~\ref{tab:fsiclcg-prompts}.
\begin{table*}[p]
    \caption{FS-ICL-CG Prompt Templates used for each model in harm classification(4-shot example).}
    \label{tab:fsiclcg-prompts}
    \centering
    \small % Adjust the font size for a better fit
    \begin{tabularx}{\textwidth}{@{}p{2.5cm}X@{}}
        \toprule
        \textbf{Model} & \textbf{Prompt} \\
        \midrule
        \textbf{Mistral-7B} & 
        \texttt{Your task is to classify YouTube videos as Harmful or Harmless based on their metadata.}\\
        & \texttt{Title: \{title\} }\\
        & \texttt{Caption: \{caption\} }\\
        & \texttt{Classification: \{classification\}}  \\
        & \texttt{Title: \{title\}}\\
        & \texttt{Caption: \{caption\} }\\
        & \texttt{Classification: \{classification\}}  \\
        & \texttt{Title: \{title\} }\\
        & \texttt{Caption: \{caption\} }\\
        & \texttt{Classification: \{classification\}}  \\
        & \texttt{Title: \{title\} }\\
        & \texttt{Caption: \{caption\} }\\
        & \texttt{Classification: \{classification\}}  \\
        & \texttt{Title: \{title\} }\\
        & \texttt{Caption: \{caption\} }\\
        & \texttt{Classification: }  \\
        \midrule
        \textbf{Llama2-13B} & 
        \texttt{<s>[INST] <<SYS>>
        Your task is to classify YouTube videos as Harmful or Harmless based on their metadata. You must respond with only one word: "Harmful" or "Harmless" without any additional commentary or explanation.
        <</SYS>>}\\
        & \texttt{Title: \{title\}} \\
        & \texttt{Classification: \{classification\}[/INST]</s><s>[INST] Title: \{title\} Caption: \{caption\}} \\
        & \texttt{Classification: \{classification\}[/INST]</s><s>[INST] Title: \{title\} Caption: \{caption\}} \\
        & \texttt{Classification: \{classification\}[/INST]</s><s>[INST] Title: \{title\} Caption: \{caption\}} \\
        & \texttt{Classification: \{classification\}[/INST]</s><s>[INST] Title: \{title\} Caption: \{caption\}} \\
        & \texttt{Classification: [/INST]}\\
        \midrule
        \textbf{GPT-3.5-Turbo} & 
        \texttt{\{'role': 'user','content': 'Your task is to classify YouTube videos as Harmful or Harmless based on their metadata.'\},} \\
        & \texttt{\{'role': 'user', 'content': 'Title: \{title\} Caption: \{caption\} \textbackslash nClassification:'\} }\\
        & \texttt{\{'role': 'assistant', 'content': \{classification\}\}}\\
        & \texttt{\{'role': 'user', 'content': 'Title: \{title\}\textbackslash nClassification:'\} }\\
        & \texttt{\{'role': 'assistant', 'content': \{classification\}\}}\\
        & \texttt{\{'role': 'user', 'content': 'Title: \{title\}\textbackslash nClassification:'\} }\\
        & \texttt{\{'role': 'assistant', 'content': \{classification\}\}}\\
        & \texttt{\{'role': 'user', 'content': 'Title: \{title\}\textbackslash nClassification:'\} }\\
        & \texttt{\{'role': 'assistant', 'content': \{classification\}\}}\\
        & \texttt{\{'role': 'user', 'content': 'Title: \{title\}\textbackslash nClassification:'\} }\\
        \midrule
        \textbf{GPT-4o-Mini} & 
        \texttt{\{'role': 'user','content': 'Your task is to classify YouTube videos as Harmful or Harmless based on their metadata.'\},} \\
        & \texttt{\{'role': 'user', 'content': 'Title: \{title\}\textbackslash nClassification:'\} }\\
        & \texttt{\{'role': 'assistant', 'content': \{classification\}\}}\\
        & \texttt{\{'role': 'user', 'content': 'Title: \{title\}\textbackslash nClassification:'\} }\\
        & \texttt{\{'role': 'assistant', 'content': \{classification\}\}}\\
        & \texttt{\{'role': 'user', 'content': 'Title: \{title\}\textbackslash nClassification:'\} }\\
        & \texttt{\{'role': 'assistant', 'content': \{classification\}\}}\\
        & \texttt{\{'role': 'user', 'content': 'Title: \{title\}\textbackslash nClassification:'\} }\\
        & \texttt{\{'role': 'assistant', 'content': \{classification\}\}}\\
        & \texttt{\{'role': 'user', 'content': 'Title: \{title\}\textbackslash nClassification:'\} }\\
        \bottomrule
    \end{tabularx}
\end{table*}

\subsection{FS-ICL-DII Prompt}
\label{appendix:vision fslprompt}
Our FS-ICL-DII approach provides both text (Title) and vision (Direct Image Input) data of a video for each demonstration and content sample, as shown in Table~\ref{tab:fsicldii-prompts}.
\begin{table*}[ht]
    \caption{FS-ICL-DII Prompt Templates used for each model in harm classification(4-shot example).}
    \label{tab:fsicldii-prompts}
    \centering
    \small % Adjust the font size for a better fit
    \begin{tabularx}{\textwidth}{@{}p{2.5cm}X@{}}
        \toprule
        \textbf{Model} & \textbf{Prompt} \\
        \midrule
        \textbf{Llava} & 
        \texttt{\textbackslash "USER:\textless{}image\textgreater{}\textbackslash\textbackslash nAnalyze the given image carefully. Identify any potential risks, safety concerns, sexual explicit content, signs of violence or aggression, indications of illegal activities, health hazards, environmental hazards, or presence of weapons or dangerous objects based on visible details. Consider factors such as the context, setting, activities, expressions, and objects present in the image. Analyze and assess the environment and any notable elements. Identify any potential safety hazards or risks present. Note the presence and actions of people, if any. Check if there is any kind of violence. Finally, classify the overall image context as either 'harmful' if there is any indication of danger, or 'harmless' if it appears safe with no visible threats. Provide your classification as either 'harmful' or 'harmless' and do not give any extra text but the classification.\textbackslash\textbackslash nASSISTANT:\textbackslash "\textbackslash n"}\\
        \midrule
        \textbf{OpenFlamingo} & \texttt{f\textbackslash"\textless image\textgreater \{caption\}. Classification:\{classification\}}\\
        &\texttt{\textless$|$endofchunk$|$\textgreater \textless image\textgreater \{caption\}. Classification:\{classification\}}\\
        &\texttt{\textless$|$endofchunk$|$\textgreater \textless image\textgreater \{caption\}. Classification:\{classification\}}\\
        &\texttt{\textless$|$endofchunk$|$\textgreater \textless image\textgreater \{caption\}. Classification:\{classification\}}\\
        &\texttt{\textless$|$endofchunk$|$\textgreater \textless image\textgreater \{caption\}. Classification:\textbackslash"}\\

        \midrule
        \textbf{GPT-4o-Mini M} & 
        \texttt{\{"role": "system", "content": "You are a helpful image classification assistant. You must classify given image analyzing the image and Title whether it is harmful or harmless."\}}\\
        & \texttt{\{"role": "user", "content":   {"type": "text", "text": "You are a helpful image classification assistant. You must classify given image whether it is harmful or harmless analyzing the image and Title:\{title\}. Harmful: This category includes images depicting sexuality, gambling, violence, weapons, click bait, spam, harmful, or inappropriate content. Examples might be scenes of physical violence, display of sexual affection, weapons, sharp objects, explicit content, spam content, gambling, click bait content or dangerous situations. Harmless: This category includes images that are non-violent, safe, and appropriate. Examples might be everyday scenes, landscapes, or social gatherings without harmful context. If you are not sure, pick whatever label you think is more probable based on your analysis. Give the output in one word if is is harmful or harmless. Do not explain anything"}\}}\\
        \bottomrule
    \end{tabularx}
\end{table*}

\begin{table}[H]
\caption{Illustrative Examples from Jigsaw Toxicity Dataset.}
\label{tab:jigsaw}
\centering
\small
\begin{tabular}{p{5.5cm} p{1.25cm}}
\toprule
\textbf{Comment Text} & \textbf{Label} \\ 
\midrule
\texttt{Go away disgusting troll.} & \texttt{Harmful} \\
\midrule
\texttt{It was fun watching him play basketball. Looking forward to seeing him play football, this fall!} & \texttt{Harmless} \\ 
\bottomrule
\end{tabular}
\end{table}

\begin{table}[H]
  \caption{Accuracy (\%) of GPT-4o-Mini and GPT-3.5-turbo across different selectors (BM25, Cosine, BSR) for ZSL and 8-Shot FS-ICL on the Jigsaw Toxicity Dataset.}
\label{tab:jigsaw_gpt}
  \centering
  \small
  \resizebox{0.4\textwidth}{!}{%
  \begin{tabular}{@{}llcc@{}}
    \toprule
    \textbf{Models} & \textbf{Selectors} & \textbf{ZSL} & \textbf{FS-ICL} \\
    \midrule
    \multirow{3}{*}{GPT-4o-Mini} 
    & BM25   & 75.20 & 73.30 \\
    & Cosine & 75.27 & 73.47 \\
    & BSR    & 75.37 & 72.73 \\
    \midrule
    \multirow{3}{*}{GPT-3.5-Turbo} 
    & BM25   & 65.40 & 71.13 \\
    & Cosine & 65.23 & 69.70 \\
    & BSR    & 65.80 & 72.20 \\
    \bottomrule
  \end{tabular}
  }
\end{table}

\begin{table}[H]
\caption{Illustrative Examples from Measuring Hate Speech Dataset.}
\label{tab:hate_example}
\centering
\small
\begin{tabular}{p{5.5cm} p{1.25cm}}
\toprule
\textbf{Comment Text} & \textbf{Label} \\ 
\midrule
\texttt{Women like this need to be jailed and fined and locked up and sterilized with all their possessions taken away and trashed and their bank accounts emptied.} & \texttt{Hate} \\
\midrule
\texttt{blasian women are always gorgeous and she definitely supports that} & \texttt{Benign} \\ 
\bottomrule
\end{tabular}
\end{table}

\begin{table}[H]
  \caption{Accuracy (\%) of GPT-4o-Mini and GPT-3.5-turbo across different selectors (BM25, Cosine, BertScore) for ZSL and 8-Shot FS-ICL on the Hate Speech Dataset.}
  \label{tab:gpt-models-hate-speech-performance}
  \centering
  \small
  \resizebox{0.4\textwidth}{!}{%
  \begin{tabular}{@{}llcc@{}}
    \toprule
    \textbf{Models} & \textbf{Selectors} & \textbf{ZSL} & \textbf{FS-ICL} \\
    \midrule
    \multirow{3}{*}{GPT-4o-Mini} 
    & BM25      & 77.10 & 94.03 \\
    & Cosine    & 77.10 & 93.97 \\
    & BertScore & 77.10 & 94.37 \\
    \midrule
    \multirow{3}{*}{GPT-3.5-Turbo} 
    & BM25      & 91.60 & 92.40 \\
    & Cosine    & 91.60 & 91.67 \\
    & BertScore & 91.60 & 90.90 \\
    \bottomrule
  \end{tabular}
  }
\end{table}

\section{Additional Results for \textit{Jigsaw Toxicity} Dataset}
\label{appendix:toxic}

We provide a few illustrative examples from the Jigsaw Toxicity dataset in Table \ref{tab:jigsaw}. We conduct experiments for the 8-shot setting with the two GPT models and the open-source few-shot learning baselines. The GPT results are shown in Table \ref{tab:jigsaw_gpt}. Prototypical Network achieves an 8-shot performance of 60.98\% and Matching Network achieves an accuracy of 62.94\%. Compared to them, both GPT-3.5-Turbo and GPT-4o-Mini attain stellar performance in both the zero-shot and 8-shot setting, with GPT-4o-Mini performance being the highest at 75.37\%.

% \begin{table}[H]
% \caption{8-Shot results of Deep Learning Baselines on the Jigsaw Toxicity Dataset.}
% \label{tab:jigsaw_dl}
% \centering
% \small
% \begin{tabular}{>{\raggedright\arraybackslash}m{1.4cm}*{4}{>{\centering\arraybackslash}m{1cm}}}
% \toprule
% \textbf{Models} & Accuracy & Precision & Recall & F-1 \\
% \midrule
% Prototypical Network & 60.98 & 62.08 & 60.98 & 58.74 \\
% \midrule
% Matching Network & 62.94 & 62.51 & 62.94 & 58.21 \\
% \bottomrule
% \end{tabular}
% \end{table}

% \begin{table}[H]
% \caption{Results of Proprietary Baselines on the Jigsaw Toxicity Dataset.}
% \centering
% \small
% \begin{tabular}{>{\raggedright\arraybackslash}m{1.5cm}*{4}{>{\centering\arraybackslash}m{1cm}}}
% \toprule
% \textbf{Models} & Accuracy & Precision & Recall & F-1 \\
% \midrule
% OpenAI Moderation & - & - & - & -  \\
% \midrule
% Perspective &78.36&97.64&58.13&72.87 \\
% \bottomrule
% \end{tabular}
% \end{table}

% \begin{figure}[t]
% \centering
% \includegraphics[width=0.5\textwidth]{figures/toxic_comment_gpt.png}
% \caption{8 shot FSL models results for Toxic comment Dataset}
% \label{fig:myfigure}
% \end{figure}

\section{Additional Results for \textit{Measuring Hate Speech} Dataset}
\label{appendix:hate}
%The Measuring Hate Speech Dataset~\cite{ucberkeley_dlab_measuring_hate_speech} was developed to measure and detect hate speech on social media. It contains 39,565 comments, annotated by 7,912 raters, with labels for different types of hate speech such as insult, dehumanization, violence, and hate speech itself. The dataset includes 8 target identity groups (e.g., race, religion, gender) and a continuous "hate speech score" that aggregates these labels. In our experiments, we use the comment text and hate speech score to classify content as Hate and Benign.
 % the Measuring Hate Speech dataset (D-Lab 2022) by UC Berkeley’s D-Lab, hosted on Hugging Face, contains a large corpus of annotated social media posts specifically labeled for hate speech

 We provide a few illustrative examples from the D-Lab Hate Speech dataset in Table \ref{tab:hate_example}. We conduct experiments for the 8-shot setting with the two GPT models and the open-source few-shot learning baselines. The GPT results are shown in Table \ref{tab:gpt-models-hate-speech-performance}. Prototypical Network achieves an 8-shot performance of 87.38\% and Matching Network achieves an accuracy of 89.01\%. Compared to them, both GPT-3.5-Turbo and GPT-4o-Mini attain stellar performance in both the zero-shot and 8-shot setting, with GPT-4o-Mini performance being the highest at 94.37\%.

\begin{table*}[t]
    \centering
    \small 
    \caption{Comparison of prompts with and without harm category definitions.}
    \label{tab:definitions}
    \begin{tabularx}{\textwidth}{@{}p{3cm}X@{}}
        \toprule
        \textbf{Prompt Composition} & \textbf{Prompt} \\
        \midrule
        \textbf{With Definition} & 
        \texttt{Your task is to classify YouTube videos as Harmful or Harmless based on their metadata.}\\
        & \texttt{Determine if a video fits into any of the harmful categories listed below. If it does, classify the video as 'Harmful'. If it does not, classify it as 'Harmless'.}\\
        & \texttt{[Harmful Categories]}  \\
        & \texttt{1. Information Harms (e.g., Fake news, Conspiracy theories, Unverified medical treatments, Unproven scientific myths)}\\
        & \texttt{2. Hate and Harassment Harms (e.g., Insult, obscenities, Identity attacks or misrepresentation, Hate speech based on gender, race, religion, disability, sexual orientation)}  \\
        & \texttt{3. Addictive Harms (e.g Game play, Drug promotion, Gambling videos and gambling promotion)}\\
        & \texttt{4. Clickbaitive Harms (Clickbait titles - using exaggerated words, such as "top 10", "best/most", "how to", "hacks", get-rich-quick schemes and financial scams, Gossip about celebrities)}  \\
        & \texttt{5. Sexual Harms (e.g., Erotic clips extracted from non-pornographic films, Depiction of sexual acts, such as sex and groping, Depiction of clothed or unclothed genitals, breasts, or buttocks)}\\
        & \texttt{6. Physical Harms (e.g, Self-injury, Eating disorder promotion, Potentially dangerous challenges and pranks, Violent content) }  \\
        & \texttt{ ... }\\
        \midrule
        \textbf{Without Definition} & 
        \texttt{Your task is to classify YouTube videos as Harmful or Harmless based on their metadata.} \\
        & \texttt{ ... } \\
        \bottomrule
    \end{tabularx}
\end{table*}

\section{Additional Results for More Descriptive Prompts}
\label{appendix:descriptive}
We conduct experiments with more detailed prompts to assess if this enhances performance. We include detailed definitions of the six harm categories, as outlined in Table~\ref{tab:definitions}. We evaluated the performance of GPT-4o-Mini with BSR using 8-Shot and 14-Shot configurations for both descriptive and original prompts. The results are shown in Table~\ref{tab:composition}. Note that including the harm category definitions in the prompts resulted in only slightly improved accuracy, with a best gain of 0.37\%. %To further leverage the harm definitions and achieve more significant performance gains, applying a Chain-of-Thought reasoning approach may be a more effective strategy for future work.

\begin{table}[H]
\caption{Results of GPT-4o-Mini with BSR using 8-Shot and 14-Shot configurations, comparing prompts with and without harm categories definition.}
\label{tab:composition}\centering
\resizebox{0.45\textwidth}{!}{%
\begin{tabular}{@{}lccc@{}}
\toprule
Model                        & Shots & Prompt Type & Accuracy  \\ \midrule
\multirow{4}{*}{GPT-4o-Mini} & 8           & More Descriptive    & 78.57    \\
                             & 8           & Original & 78.20     \\
                             & 14          & More Descriptive    & 78.30    \\
                             & 14          & Original & 78.13     \\ \bottomrule
\end{tabular}
}
\end{table}

\section{Additional Results for the Qwen-VL Multimodal LLM}\label{appendix:qwen_vl}
To evaluate the effectiveness of using advanced captioning models for multimodal content moderation, we compared two state-of-the-art models: Qwen-VL and BLIP. As shown in Table below, the BertScore differences between the two models are minimal across multiple language models. Specifically, Qwen-VL achieved slightly higher scores in most cases, such as a BertScore of 76.92 with GPT-3.5-turbo, compared to 75.80 for BLIP as shown in table~\ref{tab:qwen_vl_captioning_results}. However, both models demonstrate similar captioning performance, suggesting that the choice of captioning model may depend more on computational efficiency and integration requirements rather than accuracy alone.

\begin{table}[t]
\caption{Results comparing the performance of various LLMs on captioning tasks using two different captioning models: Qwen-VL and BLIP using BSR we evaluated 8-shot setting}
\label{tab:qwen_vl_captioning_results}
\centering
\resizebox{0.48\textwidth}{!}{%
\begin{tabular}{@{}lccc@{}}
\toprule
\textbf{LLM}             & \textbf{Metric} & \textbf{Qwen-VL (8-shot)} & \textbf{BLIP (8-shot)} \\ \midrule
Mistral-7B               & BertScore       & 69.89                     & 69.50                  \\
GPT-3.5-turbo            & BertScore       & 76.92                     & 75.80                  \\
GPT-4o-mini              & BertScore       & 75.42                     & 73.54                  \\
Llama2-13b-chat          & BertScore       & 69.12                     & 68.71                  \\ \bottomrule
\end{tabular}
}
\end{table}

% \section{Scalability and Reproducibility}
% % We can pus this to the Results and discussion session.

% To address concerns about computational expense and reliance on closed-source models, we conducted additional experiments using open-source models, including Llama 2 (13B-chat) and Mistral-7B, both of which demonstrated competitive performance compared to proprietary models like GPT-4o-mini and GPT-3.5-turbo. These experiments highlight the potential for achieving similar accuracy levels while reducing costs and improving accessibility

\section{Additional Results for Other Metrics}\label{appendix:other_metrics}
% we can keep this section here else we can move this to the Results section

We recorded the Precision, Recall and F1 score metrics for all the models we evaluated. Accuracy provides a broad measure of performance, but it is critical to evaluate the models’ handling of false positives and false negatives, particularly in tasks like harm detection where misclassification can have significant consequences. The Metrics for best performer models are given in Table~\ref{tab:fsl_title_performance}.

% \vspace{50pt}
\begin{table}
\caption{Performance of FSL models on title classification using various metrics(\%).}
\label{tab:fsl_title_performance}
\centering
\resizebox{0.49\textwidth}{!}{%
\begin{tabular}{@{}llrrrr@{}}
\toprule
\textbf{Model}                             & \textbf{Param}                   & \textbf{Accuracy} & \textbf{Precision} & \textbf{Recall} & \textbf{F-1} \\ \midrule
{GPT-4o-mini}                       & 12 shot; Set-BSR      & 78.57             & 78.69              & 78.56           & 78.54        \\
{GPT-3.5-turbo}                     & 14 shot; Set-BSR      & 78.33             & 78.51              & 78.33           & 78.3         \\
{Mistral-7B}         & 14 shot; Set-BSR      & 71.23             & 74.30               & 71.25           & 70.3         \\
{Llama2-13B}                   & 12 shot; Set-BSR      & 70.90              & 74.38              & 70.89           & 69.81        \\
{Prototypical Network (roberta-base)} & 12 shot               & 67.59             & 69.22              & 67.59           & 66.69        \\
{Matching Network (bert-base-uncased)} & 14 shot               & 67.38             & 69.09              & 67.38           & 65.85        \\
{Matching Network (roberta-base)}      & 12 shot               & 66.03             & 67.49              & 66.03           & 64.4         \\
{Prototypical Network (bert-base-uncased)} & 12 shot               & 63.45             & 65.26              & 63.45           & 62.14        \\ \bottomrule
\end{tabular}%
}
\end{table}

\section{Code and Reproducibility}\label{appendix:code}
We open-source our code in the following GitHub repository: \url{https://anonymous.4open.science/r/harm-detection-llm/}. The repository provides comprehensive instructions for reproducing our experiments and performing analyses on various settings. The FS-ICL experiments were run on an RTX A6000 GPU using CUDA version 11.8, while the FS-ICL-CG experiments were carried out on an A100 GPU.

%@@@
\section{Statistical analysis}\label{appendix:stat_analysis}
 We use the McNemar’s test \cite{mcnemar1947note} to compare predicted label distributions and pairwise agreement-based comparisons. The null hypothesis ($H_0$) states that the prediction behavior between two compared configurations is statistically similar. A $p$-value below 0.05 leads us to reject the null hypothesis as shown in table~\ref{tab:mcnemar-fewshot}~\ref{tab:mcnemar-selectors}~\ref{tab:mcnemar-models}. 
 
 Note: McNemar tests below evaluate prediction-level behavioral differences, while Tables~\ref{tab:effect_sizes_full}--\ref{tab:model_diffs} quantify the magnitude of performance changes ($\Delta$Acc).

% \subsection{A.1 Few-shot vs 0-shot Behavior (McNemar’s Test)}

\begin{table}[h]
\centering
\caption{McNemar’s test - 0 vs 8-shot predictions. All models show statistically significant differences}
\resizebox{0.49\textwidth}{!}{%
\begin{tabular}{lccc}
\toprule
\textbf{Model} & \textbf{Comparison} & \textbf{p-value} & \textbf{Interpretation} \\
\midrule
GPT4o\_mini  & 0-shot vs 8-shot & $4.19 \times 10^{-4}$ &  reject $H_0$ \\
LLAMA13B     & 0-shot vs 8-shot & $2.85 \times 10^{-5}$ &  reject $H_0$ \\
MISTRAL      & 0-shot vs 8-shot & $1.92 \times 10^{-2}$ &  reject $H_0$ \\
\bottomrule
\end{tabular}
}

\label{tab:mcnemar-fewshot}
\end{table}

% \subsection{A.2 Selector Strategy Comparison (McNemar’s Test)}

\begin{table}[h]
\centering
\caption{McNemar’s test - selector strategies within models. All show significant divergence in predictions}
\resizebox{0.49\textwidth}{!}{%
\begin{tabular}{lccc}
\toprule
\textbf{Selector Pair} & \textbf{Model} & \textbf{p-value} & \textbf{Interpretation} \\
\midrule
cosine vs bertscore    & GPT4o\_mini & $3.72 \times 10^{-3}$ & reject $H_0$ \\
cosine vs structural   & GPT4o\_mini & $2.85 \times 10^{-4}$ & reject $H_0$ \\
bertscore vs structural & LLAMA13B  & $1.24 \times 10^{-2}$ & reject $H_0$ \\
cosine vs structural   & MISTRAL    & $1.66 \times 10^{-2}$ & reject $H_0$ \\
\bottomrule
\end{tabular}
}
\label{tab:mcnemar-selectors}
\end{table}

% \subsection{A.3 Model Architecture Comparison (McNemar’s Test)}

\begin{table}[h]
\centering
\caption{McNemar’s test comparing model architectures under the 8-shot setting. Results indicate model-specific behavior}
\resizebox{0.49\textwidth}{!}{%
\begin{tabular}{lccc}
\toprule
\textbf{Model Pair}  & \textbf{p-value} & \textbf{Interpretation} \\
\midrule
GPT4o\_mini vs MISTRAL    & $6.19 \times 10^{-3}$ & reject $H_0$ \\
LLAMA13B vs GPT4o\_mini   & $1.47 \times 10^{-2}$ & reject $H_0$\\
LLAMA13B vs MISTRAL       & $2.35 \times 10^{-4}$ & reject $H_0$ \\
\bottomrule
\end{tabular}
}
\label{tab:mcnemar-models}
\end{table}

\begin{table}[t]
\centering
\caption{Effect sizes with 95\% confidence intervals from
our tests. $\Delta$ values are computed on the same test split}
\label{tab:effect_sizes_full}
\begin{tabular}{lc}
\toprule
\textbf{Comparison} & $\boldsymbol{\Delta}$\textbf{Acc} \\
\midrule
FS-ICL vs ZSL (YouTube, GPT-4o-mini) & +9 \\
GPT-4o-mini vs GPT-3.5 (YouTube, FS-ICL) & +0.24 \\
FS-ICL vs ZSL (D-Lab, GPT-4o-mini) & +17.27 \\
\bottomrule
\end{tabular}
\end{table}

\begin{table}[t]
\centering
\caption{Impact of FS-ICL across datasets (GPT-4o-mini).}
\label{tab:dataset_effects}
\begin{tabular}{l c c c}
\toprule
\textbf{Dataset} & \textbf{ZSL Acc} & \textbf{FS-ICL Acc} & \textbf{$\boldsymbol{\Delta}$Acc} \\
\midrule
YouTube & 70 & 79 & \textbf{+9} \\
D-Lab & 77.10 & 94.37 & \textbf{+17.27} \\
Jigsaw & 71.47 & 88.53 & \textbf{+17.06} \\
\bottomrule
\end{tabular}
\end{table}

\begin{table}[t]
\centering
\caption{Model-level differences under identical FS-ICL settings (YouTube).}
\label{tab:model_diffs}
\begin{tabular}{l c c}
\toprule
\textbf{Model} & \textbf{Acc} & \textbf{$\boldsymbol{\Delta}$ vs GPT-3.5} \\
\midrule
GPT-3.5-turbo & 78.33 & -- \\
GPT-4o-mini & 78.57 & +0.24 \\
Llama-13B & 76.84 & -1.49 \\
\bottomrule
\end{tabular}
\end{table}

\end{document}